%% file: manuscript.tex
\pdfoutput=1

\documentclass{article} %
\usepackage[final]{colm2026_conference}

\usepackage{microtype}
\usepackage{hyperref}
\usepackage{url}
\usepackage{amsmath}
\usepackage{booktabs}
\usepackage{graphicx}
\usepackage{multirow}
\usepackage{amssymb}
\usepackage{tcolorbox}
\usepackage{cleveref}
\newcommand{\githublogo}{\raisebox{-0.15ex}{\includegraphics[height=0.95em]{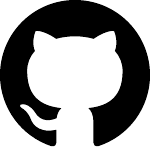}}}

\usepackage{lineno}

\definecolor{darkblue}{rgb}{0, 0, 0.5}
\hypersetup{colorlinks=true, citecolor=darkblue, linkcolor=darkblue, urlcolor=darkblue}

\title{SkillFlow: Scalable and Efficient Agent Skill Retrieval System}

\author{Fangzhou Li\textsuperscript{1,2,$\ast$}, Pagkratios Tagkopoulos\textsuperscript{1,2,3,$\ast$}, Ilias Tagkopoulos\textsuperscript{1,2,$\dagger$} \\
  \textsuperscript{1}University of California, Davis, \\
  \textsuperscript{2}USDA/NSF AI Institute for Next Generation Food Systems, \\
  \textsuperscript{3}Process Integration and Predictive Analytics
}

\begin{document}

\ifcolmsubmission
\linenumbers
\fi

\maketitle

\begingroup
\renewcommand{\thefootnote}{}%
\footnotetext{$\ast$ Equal contribution.}%
\footnotetext{$\dagger$ Correspondence to: \texttt{itagkopoulos@ucdavis.edu}}%
\endgroup

\begin{abstract}
AI agents can extend their capabilities at inference time by loading reusable skills into context, yet equipping an agent with too many skills, particularly irrelevant ones, degrades performance.
As community-driven skill repositories grow, agents need a way to selectively retrieve only the most relevant skills from a large library.
We present \textit{SkillFlow}, the first open, multi-stage retrieval system for agent skill discovery that frames skill acquisition as an information retrieval problem over a corpus of {\raise.17ex\hbox{$\scriptstyle\sim$}}35K community-contributed SKILL.md definitions indexed from GitHub.
The pipeline progressively narrows a large candidate set through four stages (dense retrieval, two rounds of cross-encoder reranking, and LLM-based selection), balancing recall and precision at each stage.
We evaluate SkillFlow on two coding benchmarks: SkillsBench, a benchmark of 87 tasks and 229 matched skills; and Terminal-Bench, a benchmark that provides only 89 tasks, and no matched skills. On SkillsBench, SkillFlow-retrieved skills raise Pass@1 from 9.2\% to 16.4\% (+78.3\%, $p_{\text{adj}} = 3.64 \times 10^{-2}$), reaching 84.1\% of the oracle ceiling, while on Terminal-Bench, agents readily use the retrieved skills (70.1\% use rate) yet show no performance gain, revealing that retrieval alone is insufficient when the corpus lacks high-quality, executable skills for the target domain.
SkillFlow demonstrates that framing skill acquisition as an information retrieval task is an effective strategy, and that the practical impact of skill-augmented agents hinges on corpus coverage and skill quality, particularly the density of runnable code and bundled artifacts.
(\githublogo~\url{https://github.com/IBPA/skill-flow})
\end{abstract}

\section{Introduction}
\label{sec:introduction}

AI agents (systems that perceive their environment and act autonomously to achieve specific goals \citep{wooldridge_intelligent_1995,russell_artificial_2016,carbonell_is_1997}) have been a long-standing pursuit, from ancient automata \citep{de_solla_price_automata_1964} to the intelligent agent programs of the 1990s and today's large language model (LLM) based agents \citep{durante_agent_2024}.
LLMs have dramatically expanded what agents can do, enabling them to reason over complex environments, generate and execute code, and solve multi-step tasks that were previously out of reach.

Despite these advances, agents today are typically equipped with a fixed set of capabilities, and the question of how they acquire and integrate new ones remains largely open.
Agent Skills, introduced by Anthropic \citep{noauthor_overview_nodate}, handles this by defining a standardized format (SKILL.md) for encoding reusable procedural knowledge.
The convention has been widely adopted by the open-source community, with a rapidly growing number of skill definitions shared across repositories on GitHub.

However, agents must load skills into their context at inference time, and equipping them with too many skills, particularly irrelevant ones, degrades both accuracy and efficiency \citep{li_skillsbench_2026,engineer_agentsmd_2026}.
Simply pre-loading all available skills is therefore impractical. What is needed is a mechanism that can selectively retrieve only the most relevant skills from a large and growing library.

\begin{figure}[t]
  \centering
  \includegraphics[width=\columnwidth]{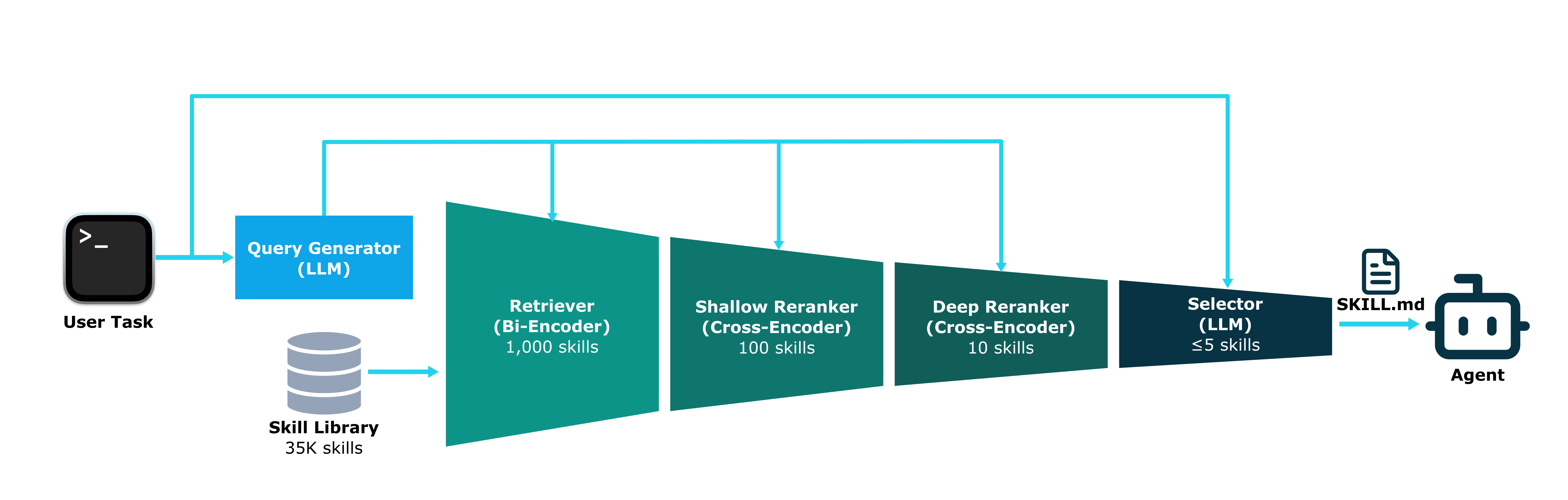}
  \caption{Overview of the SkillFlow pipeline.
  A user task is passed to an agent, which invokes SkillFlow to retrieve relevant skills from a library of 35K community-contributed skill definitions.
  The pipeline progressively narrows the candidate set through four stages: a bi-encoder retriever (1K skill candidates), a shallow cross-encoder reranker (100), a deep cross-encoder reranker (10), and an LLM-based selector ($\leq$5 skills).
  Retrieved skills are returned to the agent to augment task execution.}
  \label{fig:overview}
\end{figure}

In this work, we address this retrieval challenge and present \textit{SkillFlow}, a multi-stage pipeline that, given a task description, efficiently identifies and retrieves relevant skills from a large library of community-contributed skill definitions.
The pipeline consists of four stages: dense retrieval, shallow reranking, deep reranking, and final selection (\Cref{fig:overview}); each progressively refining the candidate set to balance recall and precision.
We evaluate SkillFlow on two coding benchmarks: on SkillsBench, retrieved skills raise Pass@1 from 9.2\% to 16.4\% ($+78.3\%$, $p_{\text{adj}} = 3.64 \times 10^{-2}$), reaching 84.1\% of the oracle ceiling.
On Terminal-Bench, however, agents readily adopt retrieved skills (70.1\% use rate) yet show no performance gain, revealing that retrieval alone is insufficient when the corpus lacks high-quality skills for the target domain.
This contrast suggests that the primary factor limiting skill-augmented agents is not retrieval but the quality and coverage of the underlying skill library.

Our contributions are as follows:
\begin{enumerate}\setlength{\itemsep}{2pt}\setlength{\parskip}{0pt}\setlength{\parsep}{0pt}
   \item We frame agent skill acquisition as information retrieval over a large, real-world corpus of community-contributed SKILL.md definitions and, to our knowledge, are the first to study skill retrieval at this scale. We build a standard multi-stage cascade (dense retrieval $\rightarrow$ shallow reranking $\rightarrow$ deep reranking $\rightarrow$ selection) for this setting and claim no novel architecture, using a standard cascade by design, so the results reflect skill retrieval itself rather than bespoke choices.
   \item We construct an openly available skill library by indexing {\raise.17ex\hbox{$\scriptstyle\sim$}}35K community-contributed SKILL.md files from GitHub, providing a reusable resource for skill-augmented agents.
   \item We evaluate SkillFlow on two benchmarks and, through a contrasting null result on Terminal-Bench, demonstrate that skill library quality, and not retrieval, is the primary bottleneck for skill-augmented agents. A structural comparison of 229 oracle skills against {\raise.17ex\hbox{$\scriptstyle\sim$}}35K community-authored skills identifies the gap precisely: oracle skills contain significantly more executable code ($p_{\text{adj}} = 3.38 \times 10^{-12}$) and are roughly 2.5 times more likely to bundle runnable scripts ($p_{\text{adj}} = 5.57 \times 10^{-18}$), while length and documentation volume do not differ. These findings translate directly into measurable authoring guidelines about skill contribution and retrieval.
\end{enumerate}

\section{Related Work}
\label{sec:related_work}

\paragraph{Skill Acquisition and Tool Discovery.}
LLM agents increasingly leverage external capabilities at inference time.
VOYAGER \citep{wang_voyager_2023} synthesizes and retrieves skills via embedding similarity, but its library is self-generated rather than drawn from a shared corpus.
Tool-calling approaches (Toolformer \citep{schick2023toolformer}, Gorilla \citep{patil2024gorilla}, ToolLLM \citep{qin2024toolllm}, and \citet{chen-etal-2025-enhancing}) retrieve atomic API signatures, whereas SkillFlow retrieves higher-level procedural knowledge: structured SKILL.md bundles that may include multi-step instructions, executable scripts, and contextual references.
Agent Skills \citep{noauthor_overview_nodate} formalized this bundle format, and a rapidly growing ecosystem now hosts tens of thousands of community-contributed definitions.
Despite this growth, existing skill-augmented agents assume a small, manually curated skill set or rely on brute-force enumeration; scalable retrieval over a large, heterogeneous skill library has not been addressed.

\paragraph{Multi-Stage Retrieval and Reranking.}
Retrieval-augmented generation \citep{lewis2021retrievalaugmentedgenerationknowledgeintensivenlp} has evolved from sparse methods like BM25 \citep{robertson2009probabilistic} to dense bi-encoders \citep{karpukhin-etal-2020-dense}, with cross-encoder rerankers trading throughput for accuracy \citep{nogueira2020passagererankingbert, nogueira2019multistagedocumentrankingbert}.
SkillFlow adopts this multi-stage architecture but operates over structured skill definitions (documents mixing natural-language descriptions with YAML metadata and embedded code) rather than homogeneous text passages, and uses a two-tier cross-encoder cascade to keep latency practical for interactive agent workflows.

\section{SkillFlow}
\label{sec:method}

Given a task description $t$ and a skill library $\mathcal{S}$ of $N$ skills, the goal is to return a small subset $\mathcal{S}^* \subset \mathcal{S}$ of the most relevant skills under a latency budget compatible with interactive agent use.
SkillFlow tackles this by progressively narrowing a large candidate set through four stages (dense retrieval, shallow reranking, deep reranking, and final selection), each refining the candidate skill set $\mathcal{C}_l$ (where $l$ denotes the stage) to balance recall and precision.
Each stage $l$ retains the top $k_l$ candidates from the previous stage's output, with $k_1 \gg k_2 \gg k_3 \geq |\mathcal{S}^*|$. Specific values are tuned via ablation experiments (Appendices~\ref{sec:retriever_details}, \ref{sec:reranker_details}, and \ref{sec:selector_details}).

\paragraph{Query Generation.}
For a given task $t$, we use an LLM to generate natural language queries that decompose the task into its constituent technologies, tools, and domains.
Each query targets a different abstraction layer (core problem domain, programming language or framework, specific libraries, and supporting tools), so that the set collectively covers the skills an agent would need.
Each pipeline stage $l$ uses a query set $\mathcal{Q}^{(l)} = \{q^{(l)}_1, \ldots, q^{(l)}_{M_l}\}$ of $M_l$ queries; when $M_l > 1$, per-query scores are aggregated by mean, max, or reciprocal rank fusion.
Full prompt templates and examples are provided in the online appendix (Appendix~\ref{sec:query_generation_details}).

\paragraph{Dense Retrieval.}
Given a skill library $\mathcal{S} = \{s_1, s_2, \ldots, s_N\}$ (\Cref{sec:skill_library}), each skill description is precomputed into a $d$-dimensional dense embedding $\mathbf{e}_i = \text{Enc}(s_i) \in \mathbb{R}^d$ using a bi-encoder, where $d$ is determined by the encoder model.
The first stage takes each query $q^{(1)}_j \in \mathcal{Q}^{(1)}$ and retrieves candidate skills by computing the cosine similarity between the query embedding $\mathbf{e}_{q_j} = \text{Enc}(q^{(1)}_j)$ and each skill embedding $\mathbf{e}_i$:
\begin{equation}
  \text{sim}(q^{(1)}_j, s_i) = \frac{\mathbf{e}_{q_j} \cdot \mathbf{e}_i}{\|\mathbf{e}_{q_j}\| \, \|\mathbf{e}_i\|}.
\end{equation}
This stage prioritizes recall, ensuring that a broad set of potentially relevant skills is retrieved for further processing.
The per-query result lists are then aggregated and deduplicated to form the first candidate set $\mathcal{C}_1$ of at most $k_1$ skills.
The aggregation strategy (e.g., union of top results per query, score fusion) is configurable (Appendix~\ref{sec:retriever_details}).

\paragraph{Shallow Reranking.}
We apply a lightweight cross-encoder model $\text{CE}_{\text{shallow}}$ to rerank the candidate skills in $\mathcal{C}_1$.
For each query $q^{(2)}_j \in \mathcal{Q}^{(2)}$ and candidate skill, the model concatenates the query with the skill content truncated to $L_{\text{shallow}}$ characters:
\begin{equation}
  r_{\text{shallow}}(q^{(2)}_j, s_i) = \text{CE}_{\text{shallow}}\!\bigl([q^{(2)}_j;\, \operatorname{trunc}(s_i, L_{\text{shallow}})]\bigr).
\end{equation}
The top-$k_2$ candidates are retained:
\begin{equation}
  \mathcal{C}_2 = \operatorname{top\text{-}k_2}\!\bigl(\mathcal{C}_1,\, r_{\text{shallow}}\bigr).
\end{equation}

\paragraph{Deep Reranking.}
Scoring all $|\mathcal{C}_1|$ candidates against full skill content is prohibitively expensive. The shallow reranker first reduces the candidate set, making full-content scoring tractable on the smaller set $\mathcal{C}_2$.
The deep reranking stage employs a heavier cross-encoder $\text{CE}_{\text{deep}}$ that attends to skill content truncated at a much larger limit $L_{\text{deep}} \gg L_{\text{shallow}}$, enabling a more thorough evaluation of relevance:
\begin{equation}
  r_{\text{deep}}(q^{(3)}_j, s_i) = \text{CE}_{\text{deep}}\!\bigl([q^{(3)}_j;\, \operatorname{trunc}(s_i, L_{\text{deep}})]\bigr).
\end{equation}
The top-$k_3$ candidates are retained:
\begin{equation}
  \mathcal{C}_3 = \operatorname{top\text{-}k_3}\!\bigl(\mathcal{C}_2,\, r_{\text{deep}}\bigr).
\end{equation}

\paragraph{Final Selection.}
The final stage uses an LLM-based filter $f_{\text{select}}$ that evaluates each candidate in $\mathcal{C}_3$ for \emph{relevancy}, whether the skill is topically relevant to the task $t$, and retains only those that pass:
\begin{equation}
  \mathcal{S}^* = f_{\text{select}}(t,\, \mathcal{C}_3), \quad |\mathcal{S}^*| \leq |\mathcal{C}_3|.
\end{equation}
Unlike the previous stages, which output a fixed number of candidates, the selector outputs a variable number of skills $|\mathcal{S}^*|$, compacting the injected set for the downstream agent.
Optionally, a second \emph{specificity filter} can be chained to further assess whether surviving skills provide actionable, domain-specific guidance beyond what the agent can derive from its training data (Appendix~\ref{sec:selector_details}).

\section{Experiments and Results}

\subsection{Datasets}

\paragraph{Skill Library.}
\label{sec:skill_library}
We construct a skill library by collecting skills from SkillsMP, an open marketplace of community-contributed agent skills.
Skills are enumerated via the SkillsMP API. For each skill, the associated GitHub repository is downloaded and only those containing a valid SKILL.md file are retained.
Of these, 35{,}129 contain a top-level SKILL.md and form the final indexed library; the remainder download successfully but ship no SKILL.md and are discarded.
Each skill entry includes a name, a natural language description, and associated metadata.
The full skill descriptions are embedded using an encoder-based embedding model to support dense retrieval.
Details of the crawling pipeline and corpus statistics are provided in the online appendix (Appendix~\ref{sec:skill_library_construction}).

\paragraph{SkillsBench.}
SkillsBench \citep{li_skillsbench_2026} is a benchmark of 87 tasks, each paired with one or more human-authored \emph{oracle skills}, curated skill definitions known to be relevant to the task (229 total).
Skills are co-authored alongside tasks by domain experts and validated through automated checks and human review, establishing topical relevance (the skills address the right domain and techniques), though not guaranteed benefit on every task-agent combination.
We treat these pairings as retrieval ground truth by injecting the oracle skills into the Skill Library and measuring whether SkillFlow can recover them (Experiments 1, 2, and 3).

\paragraph{Terminal-Bench.}
Terminal-Bench \citep{merrill2026terminalbench} is a benchmark of terminal-based coding tasks.
It consists of 89 tasks that test an agent's ability to perform operations in a terminal environment.
Unlike SkillsBench, Terminal-Bench does not provide oracle skill assignments, so we use it for end-to-end evaluation only (Experiments 1 and 2).

\subsection{Settings}

For query generation (Appendix~\ref{sec:query_generation_details}), we use OpenAI gpt-4o-mini \citep{hurst2024gpt}.
The dense retriever generates $M_1{=}5$ queries per task and uses Hugging Face BAAI/bge-base-en-v1.5 \citep{xiao2024c} as the bi-encoder, retrieving the top $k_1 = 1000$ candidates in total.
The shallow reranker uses SBERT cross-encoder/ms-marco-MiniLM-L-6-v2 \citep{reimers2019sentence} as a cross-encoder with $L_{\text{shallow}} = 512$ characters and a single query ($M_2{=}1$), retaining the top $k_2 = 100$ candidates.
The deep reranker uses Hugging Face BAAI/bge-reranker-v2-m3 \citep{chen2024m3} as a cross-encoder with $L_{\text{deep}} = 4096$ characters, also with $M_3{=}1$, retaining the top $k_3 = 10$ candidates.
The final selector uses OpenAI gpt-4o-mini \citep{hurst2024gpt} to select up to 5 skills.
For end-to-end benchmark evaluation, we use Codex CLI with OpenAI gpt-5-mini \citep{singh2025openai} at medium reasoning effort as the primary agent, and Claude Code with Claude Haiku 4.5 as a second agent on SkillsBench to test whether the gains transfer across harnesses and vendors (\Cref{sec:experiment_benchmark}).
We report results using the best-performing pipeline configuration. Additional experiments on key hyperparameters (e.g., $k$ values, query count $M$, and model choices) are provided in the online appendix.
For the benchmark harness, we rely on Harbor \citep{Harbor_Framework_Team_Harbor_A_framework_2026} with the Daytona Sandboxes cloud service.
Except proprietary OpenAI models, all language model components in SkillFlow are run on an Nvidia RTX A5000 GPU.

\paragraph{Metrics.}
Each condition is run 3 times.
For end-to-end evaluation (Experiments 1--2), we report Pass@$k$ (the probability that at least one of $k$ attempts succeeds, estimated from the 3 runs via the unbiased combinatorial estimator \citep{chen2021evaluating}) and skill use rate (the fraction of tasks where the agent uses at least one retrieved skill).
For retrieval evaluation (Experiment 3), we report recall@$k$ (fraction of oracle skills in the top $k$), precision@$k$ (fraction of top $k$ that are oracle skills), mean reciprocal rank (MRR, the reciprocal of the rank of the first oracle skill), and hit@$k$ (fraction of tasks with at least one oracle skill in the top $k$), all averaged across tasks.
All confidence intervals are 95\% bootstrap (10{,}000 resamples over tasks).
Pairwise significance is assessed via paired bootstrap tests. Reported $p_{\text{adj}}$ values are Holm-Bonferroni corrected for multiple comparisons within each benchmark.
Full statistical details are in the online appendix (Appendix~\ref{sec:statistical_methodology}).

\subsection{Experiment 1: Downstream Benchmark Performance}
\label{sec:experiment_benchmark}

To evaluate end-to-end effectiveness, we measure whether skills retrieved by SkillFlow improve agent task completion.
We benchmark on 65 SkillsBench tasks and 88 Terminal-Bench tasks after removing problematic tasks (Appendix~\ref{sec:task_exclusion}).
Both benchmarks are evaluated under four conditions: (1) no skills, (2) Vercel skills (a third-party skill retrieval API; Appendix~\ref{sec:vercel_details}), (3) SkillFlow-retrieved skills, and (4) oracle skills (SkillsBench only, as Terminal-Bench does not provide oracle assignments).

\begin{table}[t]
  \centering
  \caption{Benchmark performance comparison.
  Pass@$k$: probability of at least one success in $k$ independent attempts; Pass\textasciicircum$k$: probability that all $k$ attempts succeed.
  Oracle skills are the curated task-skill pairs provided by SkillsBench.
  $^{*}p_{\text{adj}}<0.05$, $^{**}p_{\text{adj}}<0.01$ vs.\ no skills baseline.}
  \label{tab:results}
  \input{tables/1_results}
\end{table}

\paragraph{Downstream gains depend on skill library coverage.}
\Cref{tab:results} summarizes the results.
On SkillsBench, where oracle skills exist in the library, SkillFlow raises Pass@1 from 9.2\% to 16.4\% (+78.3\%), a statistically significant improvement ($p_{\text{adj}} = 3.64 \times 10^{-2}$; Cohen's $h = 0.22$, small effect), reaching 84.1\% of the oracle ceiling (19.5\%); the trend holds at Pass@3 (29.2\% vs.\ 16.9\% baseline), while Vercel shows no significant improvement (9.7\%, $p_{\text{adj}} \geq 0.05$).
Steps and cost per task remain comparable across conditions, indicating that gains stem from skill quality rather than increased computation.
The gain replicates on a second agent: on Claude Code with Claude Haiku 4.5 (\Cref{tab:results}, three runs), the Oracle $>$ SkillFlow $>$ no-skill ordering holds with a +159\% relative gain (Pass@1 6.7\% $\to$ 17.4\%, $p_{\text{adj}} = 2.00 \times 10^{-4}$, Cohen's $h = 0.34$; oracle ceiling 30.3\%).
The selector affects only context size: injecting the deep reranker's top-10 skills directly matches full SkillFlow at equal cost (Pass@1 15.9\% vs.\ 16.4\%, paired bootstrap $p{=}0.91$; \Cref{tab:selector_ablation}).

On Terminal-Bench, however, the no-skill baseline is already strong (Pass@1 = 34.8\%, Pass@3 = 46.6\%), and no condition produces a significant difference: neither SkillFlow (31.8\%, $p_{\text{adj}} \geq 0.05$) nor Vercel (32.6\%, $p_{\text{adj}} \geq 0.05$) improve over it.
This contrast (gains where oracle skills exist, none where the library lacks coverage, and holding across two agents) suggests that skill quality, not retrieval, is the primary bottleneck.
\Cref{tab:results} also reports a SkillFlow-specific variant with additional specificity filtering. We analyze this condition in \Cref{sec:discussion}.

\begin{figure}[t]
  \centering
  \includegraphics[width=\columnwidth]{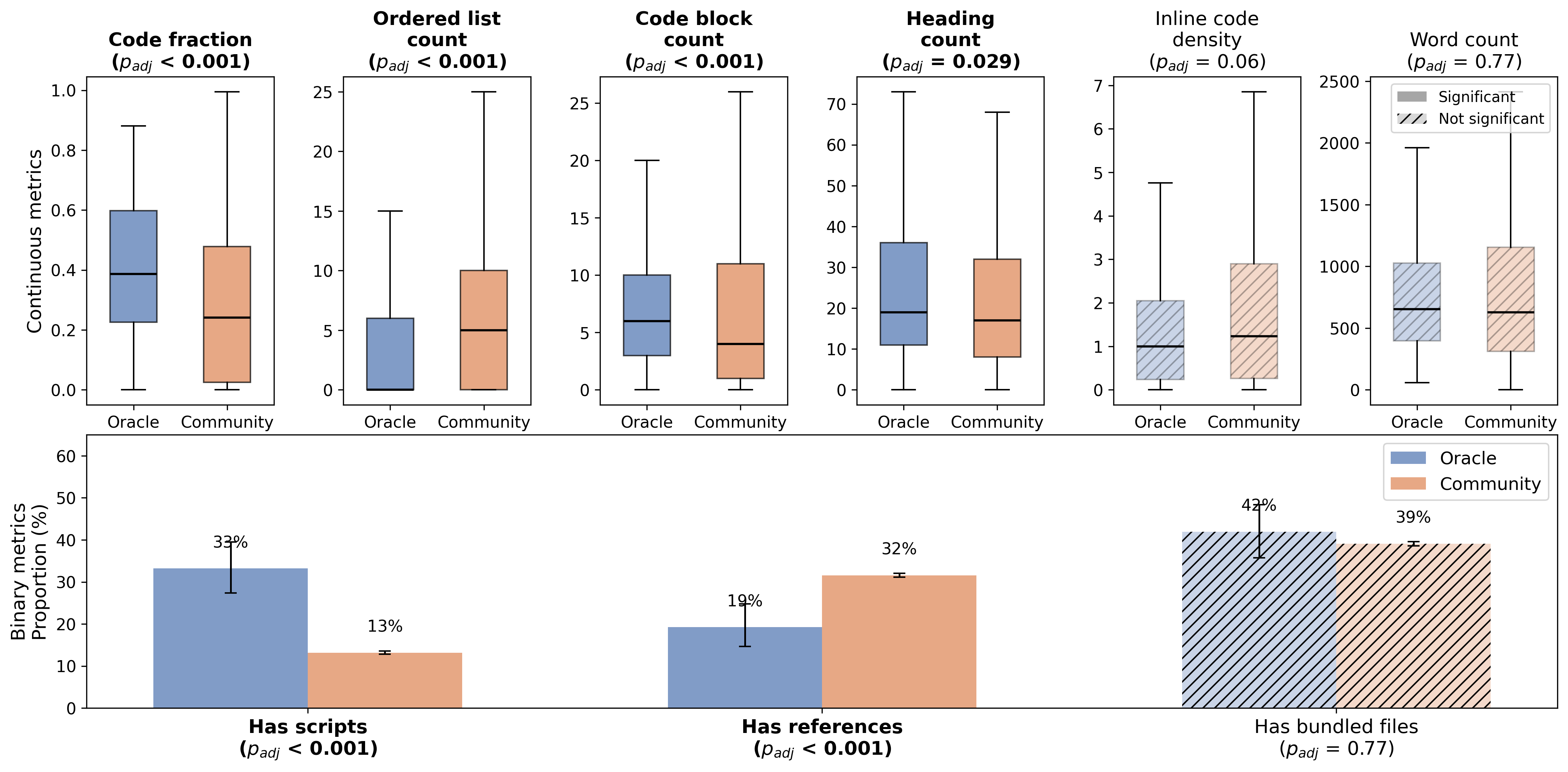}
  \caption{Structural proxy metrics: oracle ($n{=}229$) vs.\ community ($n{=}35{,}129$) skills.
  Subplots sorted by significance. $p$-values from two-sided Mann-Whitney $U$ tests, adjusted via Holm-Bonferroni.
  Hatched boxes/bars indicate non-significant differences ($p_{\text{adj}} \geq 0.05$).
  Outliers suppressed for clarity.}
  \label{fig:skill_quality_proxies}
\end{figure}

\paragraph{Characterizing the quality gap.}
We compare structural properties of 229 oracle skills against the 35{,}129 indexed community skills using automated proxy metrics (\Cref{fig:skill_quality_proxies}; two-sided Mann-Whitney $U$, Holm-Bonferroni adjusted).
Oracle skills are not significantly longer ($p_{\text{adj}} \geq 0.05$), but devote a larger fraction to fenced code blocks (0.39 vs.\ 0.24, $p_{\text{adj}} = 3.38 \times 10^{-12}$) and are roughly 2.5 times more likely to bundle executable scripts (33.2\% vs.\ 13.2\%, $p_{\text{adj}} = 5.57 \times 10^{-18}$).
Conversely, community skills favor ordered lists and reference documents, implying a more procedural, text-heavy style.
What distinguishes effective skills is not length or documentation volume, but the density of runnable code and bundled artifacts.

\subsection{Experiment 2: Skill Retrieval and Usage}
\label{sec:experiment_adoption}

To understand whether agents actually use retrieved skills, we compare SkillFlow against Vercel, a third-party skill retrieval API (Appendix~\ref{sec:vercel_details}), measuring retrieval and skill use statistics across both benchmarks.
For each task, each method retrieves skills from the library. The agent then decides which (if any) to load into its context.
On SkillsBench, oracle task--skill pairs serve as relevancy ground truth. Terminal-Bench lacks such annotations.

\begin{table}[t]
  \centering
  \caption{Skill retrieval and agent use statistics.
  Tasks Retrieved: percentage of tasks for which $\geq$1 skill was retrieved.
  Oracle Skills Retrieved: percentage of oracle skills retrieved from the library.
  Tasks Used: percentage of tasks where the agent used $\geq$1 retrieved skill.
  Vercel maintains its own proprietary skill index that does not include SkillsBench oracle skills.
  Terminal-Bench does not provide oracle skill assignments.
  $^{*}p_{\text{adj}}<0.05$; $^{**}p_{\text{adj}}<0.01$ on Tasks Used vs.\ Vercel baseline.}
  \label{tab:adoption}
  \input{tables/2_adoption}
\end{table}

\paragraph{SkillFlow achieves higher retrieval and use rates.}
\Cref{tab:adoption} reports the results.
On SkillsBench, SkillFlow retrieves 61.8\% of oracle skills from the 35K library.
This comparison is unavailable for Vercel, whose proprietary index does not include the SkillsBench oracle skills.
Agents use SkillFlow skills at a 68.8\% rate ($p_{\text{adj}} < 1 \times 10^{-6}$ vs.\ Vercel) with an average of 2.8 skills retrieved per task, approaching the oracle use rate of 69.2\%, indicating that the pipeline consistently surfaces skills the agent judges useful.
On Terminal-Bench, SkillFlow retrieves skills for 87.5\% of tasks compared to Vercel's 90.9\%, and agents use SkillFlow skills at a significantly higher rate (70.1\% vs.\ 22.3\%, $p_{\text{adj}} < 1 \times 10^{-6}$).

Even when restricted to a single top-ranked skill (matching Vercel's cardinality), SkillFlow achieves significantly higher use rates (68.8\% vs.\ 40.6\% on SkillsBench; 72.7\% vs.\ 22.3\% on Terminal-Bench; both $p_{\text{adj}} < 1 \times 10^{-6}$), confirming that the advantage stems from retrieval quality.

\subsection{Experiment 3: Stage-Level Retrieval Performance}
\label{sec:experiment_retrieval}

This experiment evaluates SkillFlow's retrieval quality at each pipeline stage.
We use the 229 oracle skills from the 87 SkillsBench tasks as retrieval ground truth and measure their overlap with SkillFlow's output at each stage.

\begin{table}[t]
  \centering
  \caption{Stage-level retrieval performance.
  Recall (R), precision (P), and mean reciprocal rank (MRR) across each stage of the SkillFlow retrieval pipeline.
  Inapplicable cells (where $k$ exceeds the stage output size) are marked with dashes.
  K output denotes the number of candidates produced by each stage.}
  \label{tab:retrieval_stages}
  \input{tables/3_retrieval_stages}
\end{table}

\paragraph{Reranking improves top-heavy ranking, while selection trades recall for precision.}
\Cref{tab:retrieval_stages} reports recall, precision, and MRR at each stage.
The dense retriever achieves R@1000 of 0.905. The shallow reranker compresses to 100 candidates while boosting R@1 by 48.9\% and MRR from 0.487 to 0.587.
The deep reranker narrows to 10 candidates with MRR of 0.634. The selector then trades recall for precision, lowering R@5 but achieving the highest P@1 (0.563), consistent with a precision-oriented compaction gate; this recall reduction does not lower downstream Pass@1, since the no-selector ablation matches full SkillFlow (\Cref{sec:experiment_benchmark}).
A stage ablation (\Cref{tab:stage_ablation}) further confirms that progressive refinement outperforms alternative first-stage methods (BM25, hybrid), with the deep reranker achieving the best R@5 (0.499) and Hit@10 (0.793); adding the selector raises MRR to 0.639 while lowering R@5 to 0.455, the same precision-for-recall trade shown in \Cref{tab:retrieval_stages}.

\begin{table}[t]
  \centering
  \caption{Stage-ablation retrieval metrics on SkillsBench (87 tasks).
  MRR, R@5, R@10, P@10, and Hit@10 (fraction of tasks with $\geq$1 oracle skill in top 10) reported.
  The first three rows compare retrieval methods. Subsequent rows add pipeline stages cumulatively.
  The selector emits at most 5 skills, so its @10 metrics are inapplicable (dashes); it raises MRR while lowering R@5.}
  \label{tab:stage_ablation}
  \input{tables/4_stage_ablation}
\end{table}

\begin{figure}[t]
  \centering
  \includegraphics[width=\columnwidth]{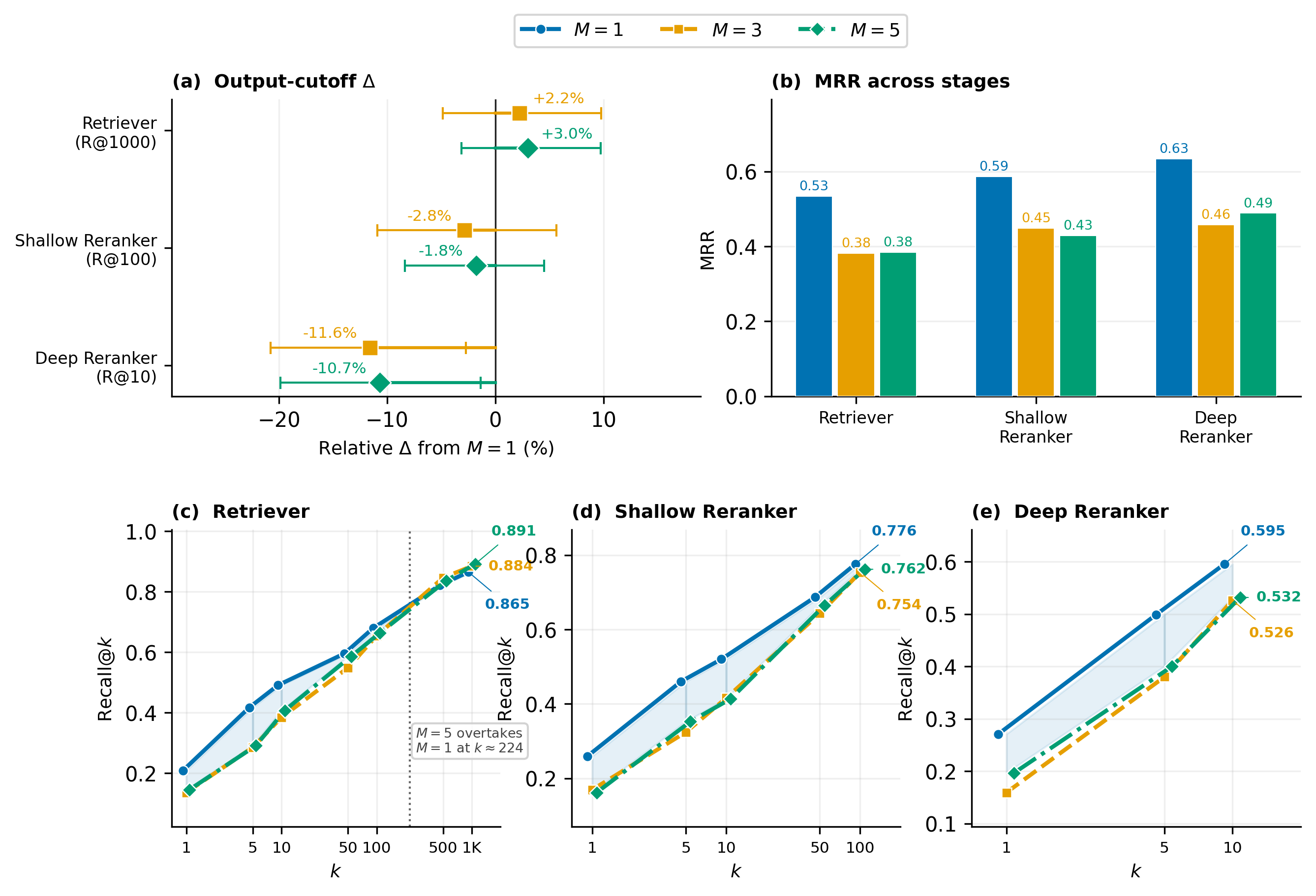}
  \caption{Multi-query impact across pipeline stages.
  (a)~Relative change in pass-through recall (R@$k_\text{output}$) from the $M{=}1$ baseline.
  (b)~MRR by stage, showing that $M{=}1$ yields the best top-heavy ranking at every stage.
  (c--e)~Recall@$k$ curves for each stage; the dashed line in~(c) marks where $M{=}5$ overtakes $M{=}1$.
  Multi-query improves retriever coverage but degrades reranker discrimination.}
  \label{fig:query_impact}
\end{figure}

\paragraph{Performance Impact of Query Generation.}
\Cref{fig:query_impact} shows the effect of the number of generated queries ($M$) on each pipeline stage.
Multi-query generation ($M{=}5$) improves retriever pass-through recall by +3.0\% by casting a wider net, but degrades reranker performance ($-$10.7\% at the deep reranker, $p_{\text{adj}} = 4.92 \times 10^{-2}$ for $M{=}3$).
$M{=}1$ achieves the highest MRR at every stage (panel~b), confirming that on small, already-filtered candidate sets, additional queries introduce noise.
The R@$k$ curves (panels~c--e) show that $M{=}5$ overtakes $M{=}1$ only beyond $k{\approx}224$ at the retriever, while $M{=}1$ dominates at all reranker cutoffs.
These results motivate our design: $M{=}5$ for retrieval to maximize recall, $M{=}1$ for reranking to maximize ranking quality.

\paragraph{Structure-aware indexing raises the recall ceiling.}
Motivated by description-only embedding beating full-content embedding (\Cref{tab:retriever_comparison}), we test a structure-aware index that encodes the YAML, prose, and code sections of each SKILL.md as three independent dense indices fused at the score level.
On a matched single-query eval, the 3-way split raises recall at every cutoff (R@1000 from 0.843 to 0.925, R@100 by up to $+13.2$pp) at a small cost in top-1 ranking, while a code-specialized encoder does not help in this natural-language-driven setting (\Cref{tab:structure_aware}).
Because this gain sits at the first-stage recall ceiling that bounds every downstream stage, the sectioned index is a promising representation; we leave its end-to-end evaluation to future work.

\subsection{Qualitative Analysis: Why Gains Are Bounded}
\label{sec:qualitative}

Three analyses explain why gains stay bounded even where SkillFlow retrieves well (full tables in the appendix).
\emph{Retrieval blind spots:} An oracle skill's value often lies in its bundled runnable scripts, not its prose: in the court-form task (\Cref{tab:case_study_2}) the oracle skill helps only because it ships executable scripts that a description-based retriever and a truncated reranker never see: the structural gap of \Cref{fig:skill_quality_proxies}.
\emph{Query-generation gap:} Some oracle skills are generic utilities the task never names; the out-of-query share rises from 4\% among top-5 oracles to 48\% among the worst-ranked (\Cref{tab:oracle_query_gap}), e.g.\ a Civilization~6 task needing the generic \texttt{hex-grid-spatial} skill, motivating artifact-aware query generation (Future directions, \Cref{sec:discussion}).
\emph{Help vs.\ hurt:} Across the SkillsBench runs, retrieved skills improve 16 of 64 tasks and regress 4 (paired bootstrap $p{=}0.02$; \Cref{tab:skill_usage}); agents open non-oracle skills only 17\% of the time versus 42\% for oracle skills, and gains concentrate where an oracle skill was recovered, motivating skill adaptation as future work.

\section{Discussion}
\label{sec:discussion}

\paragraph{SkillFlow enables scalable retrieval.}
The multi-stage pipeline efficiently trades breadth for precision: the dense retriever captures 90.5\% of oracle skills across 35K candidates, and each subsequent stage improves top-heavy precision while maintaining comparable steps and cost per task.
The full pipeline completes in ${\sim}$35 seconds per task (median), dominated by the deep reranker (${\sim}$26s); per-stage latency is reported in \Cref{tab:latency}.
The single-query configuration ($M{=}1$ for reranking) offers the best latency--quality balance, keeping end-to-end time well within typical agent task durations and making large skill libraries practical for real-time use.
This latency is largely hardware-dependent (the deep reranker runs locally on an RTX A5000) and reducible with a smaller cross-encoder or a hosted reranker API.

\begin{table}[t]
  \centering
  \caption{Per-stage wall-clock latency (seconds) on 87 SkillsBench tasks, Nvidia RTX A5000 GPU.}
  \label{tab:latency}
  \input{tables/5_latency}
\end{table}

\paragraph{Retrieval gains are bounded by what the library contains.}
On SkillsBench, where oracle skills exist, SkillFlow raises Pass@1 by 78.3\% ($p_{\text{adj}} = 3.64 \times 10^{-2}$), reaching 84.1\% of the oracle ceiling.
On Terminal-Bench, where the library lacks domain coverage, no condition differs significantly from baseline ($p_{\text{adj}} \geq 0.05$, Cohen's $h < 0.07$), confirming that the bottleneck is what is \emph{in} the library, not how we search it.
The structural quality analysis (\Cref{fig:skill_quality_proxies}) reinforces this: effective skills are distinguished by the density of runnable code and bundled artifacts, a gap that retrieval alone cannot close.
Retrieval itself is not the limiting factor: SkillFlow recovers 90.5\% of oracle skills at the retriever (\Cref{sec:experiment_retrieval}) and reaches 84.1\% of the oracle Pass@1 ceiling (\Cref{sec:experiment_benchmark}), so the low absolute ceiling reflects downstream factors (agent non-adoption, task difficulty, and variance) and is model-dependent (19.5\% on GPT-5-mini, 30.3\% on Claude Haiku 4.5), so we expect it to rise with stronger agents, a larger library, and the retrieval upgrades above.
Since Terminal-Bench has no oracle assignments, deriving validated oracle skills for such domains (SkillsBench used 105 contributors) is an open community direction, not a limitation of retrieval.

\paragraph{Filtering low-quality skills.}
We augment SkillFlow with a specificity gate, sequential relevancy and specificity LLM filters, producing the SkillFlow-specific variant (\Cref{tab:results}).
The effect is negligible on Terminal-Bench ($p_{\text{adj}} \geq 0.05$), and the reason is placement: by the specificity stage the relevancy pass has already removed the topically irrelevant candidates, leaving little to change, so the specificity filter is a cheap safety check rather than a load-bearing stage.
The remaining quality gap (code density, bundled artifacts) is structural rather than semantic, and thus not easily captured by an LLM-based filter.

\paragraph{Future directions.}
Natural extensions include expanding the library through automated extraction from repositories or agent interaction logs, peer-to-peer skill sharing, and \emph{interleaved retrieval}, retrieving skills dynamically during task execution rather than only beforehand.
On the retrieval side, \emph{artifact-aware query generation}, inspecting each task's files to enumerate implied tools and issuing a query per dependency, would reach the out-of-query oracle skills that task descriptions miss (\Cref{sec:qualitative}), and per-candidate \emph{snippet summarization} could preserve signal in a smaller reranker window; code-heavy benchmarks such as FreshStack are a natural evaluation target.
Because retrieved skills are only sometimes useful as-is, \emph{skill adaptation} (editing retrieved skills to the task) and automating high-quality skill and oracle derivation are promising ways to raise the ceiling and extend the method to sparse domains.

\paragraph{Limitations.}
We evaluate two agent models (Codex CLI with GPT-5-mini and Claude Code with Claude Haiku 4.5) across two vendors and two benchmarks; broader coverage of architectures and domains remains future work.
Our corpus audit finds only about 32\% of community skills are runnable-complete (\Cref{sec:skill_library_construction}), motivating stronger vetting before skills enter the corpus; we do not verify runtime correctness, and the quality-gap analysis relies on structural proxies and an in-context LLM judge rather than human judgments.
The skill library was constructed at a fixed point in time; a more up-to-date library would likely improve coverage.
Retrieval also sees each skill only through its description and a truncated window, so bundled scripts and long-tail dependencies stay invisible; snippet summarization and artifact-aware query generation are the natural remedies, and we leave both to future work.

\section*{Ethics Statement}
Aggregating community-contributed skills at scale raises attribution and licensing concerns.
Skills in the library are sourced from public GitHub repositories under various open-source licenses. Our crawling pipeline preserves repository metadata but does not currently enforce license-specific constraints on redistribution or modification.
Practitioners deploying SkillFlow should implement license filtering appropriate to their use case.
Additionally, a centralized skill library could be a vector for skill poisoning, malicious skills crafted to induce harmful agent behavior, motivating future work on skill provenance verification and content safety filtering.

\section*{Reproducibility Statement}
To support reproduction, we release the crawler and skill metadata/index, the retrieval code (dense retrieval, shallow and deep rerankers, and the LLM selector), the query-generation prompts and cached generated queries, the benchmark harness scripts, the task-exclusion list, and the statistical analysis scripts, together with instructions to reconstruct the {\raise.17ex\hbox{$\scriptstyle\sim$}}35K-skill corpus from its source GitHub URLs.
All artifacts are available at \url{https://github.com/IBPA/skill-flow}.
Because upstream repositories may change or be removed, we record per-skill content hashes and download timestamps so that a reconstructed corpus can be validated against the version used in our experiments.

\paragraph{Disclosure of LLM usage.}
LLMs were used for code implementation, with all code reviewed by the authors. LLMs were also used for refining paper writing with human proofreading. Experimental results, tables, figures, and references were generated by the authors manually or programmatically.

\section*{Acknowledgments}
This work was supported by the USDA-NIFA AI Institute for Next Generation Food Systems (AIFS), USDA-NIFA award number 2020-67021-32855.

\bibliography{references}

\appendix
\section{Appendix}

\subsection{Skill Library Construction}
\label{sec:skill_library_construction}

Skills were collected from SkillsMP, an open marketplace hosting over 80,259\footnote{As of 1/24/2026 in https://skillsmp.com} agent skills.
Each skill is a structured bundle containing a Markdown document (SKILL.md) with YAML frontmatter containing metadata (name, description, allowed tools) and free-form instructional content such as code examples, reference material, and tutorials.

\paragraph{Crawling Pipeline.}
The collection pipeline operated in three stages.
First, skills were enumerated via the SkillsMP API with page-based pagination. For each skill, metadata was recorded including name, description, author, tags, star count, and associated GitHub repository URL.
Second, the corresponding GitHub repository (or subdirectory) for each skill was downloaded as a ZIP archive via the GitHub API.
A configurable repository size limit of 50\,MB was enforced to exclude oversized repositories.
Downloads used rate-limited asynchronous HTTP requests (1.0\,s inter-request delay) with exponential backoff retry (1--10\,s, up to 3 attempts) for transient failures.
The pipeline supported resumable crawling via persisted sync state.
Third, only skills containing a valid SKILL.md file were retained. Skills hosted on deleted or inaccessible repositories (HTTP 404/403) were permanently skipped.

\paragraph{Corpus Statistics.}
\Cref{tab:corpus_stats} summarizes the collection results.
Of the 43,660 skills processed, 7,703 were excluded due to the 50\,MB repository size filter and 91 failed due to deleted or inaccessible repositories, yielding 35{,}866 downloaded skills, of which 35{,}129 carry a valid top-level SKILL.md and are embedded into the retrieval index.

\begin{table}[t]
  \centering
  \caption{Skill corpus collection statistics.}
  \label{tab:corpus_stats}
  \input{tables/6_corpus_stats}
\end{table}

\paragraph{Runnable-completeness audit.}
The structural analysis (\Cref{fig:skill_quality_proxies}) shows a code-density gap; here we quantify how much of the corpus is actually functional.
Structurally, 78.1\% of the 35{,}129 indexed community skills are code-bearing (containing at least a fenced code block or a bundled script), but only 13.2\% bundle an executable script (vs.\ 33.2\% of oracle skills).
To go beyond structure, an LLM judge scored 300 stratified-sampled skills in context (SKILL.md plus the bundled-file listing) for code-soundness, missing files, and purpose-alignment (\Cref{tab:functional_audit}).
By this audit, 32.3\% are \emph{runnable-complete}, shipping complete runnable code with no missing bundled files (95\% CI $[27.3, 37.8]$), while 59\% are reference-only prose and 9\% intend to ship code that is incomplete or references absent files.
At scale this is roughly 11.5K runnable-complete skills, still a substantial resource, and it quantifies the corpus-quality gap that motivates our work.
Verifying that this code produces exactly correct output would require executing every skill and is out of scope.

\begin{table}[t]
  \centering
  \caption{Runnable-completeness audit of 300 stratified-sampled community skills, scored by an in-context LLM judge.
  Runnable-complete: ships complete runnable code with no missing bundled files. 95\% Wilson confidence interval.}
  \label{tab:functional_audit}
  \input{tables/19_functional_audit}
\end{table}

\paragraph{Data Format.}
Each skill is stored as a directory containing a SKILL.md file in structured Markdown format with YAML frontmatter (name, description, allowed tools) followed by instructional content.
Skill metadata (including name, source URL, GitHub URL, author, tags, star count, content hash, and download timestamp) is persisted in a JSON index for downstream retrieval system indexing.

\subsection{Query Generation Details}
\label{sec:query_generation_details}

Given a task description, query generation produces concise natural language queries that capture the core technical skills required.
These queries serve as inputs to the downstream retrieval and reranking stages.
Two prompt variants are used depending on the pipeline stage: a \textbf{single-query prompt} (\Cref{box:single_query_prompt}), used by the reranker and deep reranker stages ($M = 1$), which produces one concise query capturing the primary technology or technique; and a \textbf{multi-query prompt} (\Cref{box:multi_query_prompt}), used by the retriever stage ($M > 1$), which produces $M$ diverse queries covering different layers of the task (core domain, language/framework, specific libraries, and supporting tools).
Both prompts use gpt-4o-mini \citep{hurst2024gpt} with a maximum output length of 200 tokens and temperature 0.0.
\Cref{tab:query_examples} shows example outputs for both variants on a single task.

\begin{figure}[t]
\begin{tcolorbox}[title=Single-Query Prompt, colback=gray!5, colframe=gray!50, fonttitle=\bfseries\small, fontupper=\small\ttfamily\raggedright]
You are a search query generator. Given a detailed task instruction for a coding agent, generate a concise search query (1-2 sentences, under 200 characters) that captures the core technical skill needed. Focus on the primary technology, tool, or technique required. Omit file paths, specific data values, and implementation details.
\end{tcolorbox}
\caption{Single-query prompt used by the reranker and deep reranker stages ($M = 1$).}

\label{box:single_query_prompt}
\end{figure}

\begin{figure}[t]
\begin{tcolorbox}[title=Multi-Query Prompt, colback=gray!5, colframe=gray!50, fonttitle=\bfseries\small, fontupper=\small\ttfamily\raggedright]
You are a search query generator for a skill retrieval system. A skill is a self-contained reference document that teaches an AI agent how to use a specific technology, tool, library, or technique.

Given a task instruction, generate \{num\_queries\} diverse search queries that would each match a different skill an agent might need. Each query should be 1-2 sentences, under 200 characters.

Guidelines for generating queries:

1. Cover different LAYERS of the task: the core problem domain, the programming language/framework, specific libraries or file formats involved, and any supporting tools or techniques.

2. Name concrete technologies: if the task mentions .xlsx files, one query should be about spreadsheet/Excel manipulation. If it involves a specific framework, name it.

3. Think about what SKILLS the agent needs, not the step-by-step procedure. Ask `what would someone search for to find a how-to guide for this part of the task?'

4. Vary abstraction levels: include both specific tool queries (e.g.\ `openpyxl spreadsheet editing') and broader domain queries (e.g.\ `data analysis with Python').

5. Do NOT generate queries about generic file I/O, writing output files, or basic Python operations; these are too broad to match any specific skill.

Respond with ONLY a JSON array of strings, no other text.
\end{tcolorbox}
\caption{Multi-query prompt used by the retriever stage ($M > 1$).}

\label{box:multi_query_prompt}
\end{figure}

\paragraph{Caching.}
Generated queries are cached to a JSON file keyed by task ID.
This ensures identical queries are reused across experiment variants; for example, when comparing aggregation strategies, only the aggregation changes while the queries remain fixed.

\paragraph{Retry and Fallback.}
If multi-query JSON parsing fails, the system retries up to 2 times, increasing temperature by 0.3 per attempt.
If all retries fail, it falls back to $M$ independent single-query calls with incrementally increasing temperature (+0.3 per call) to encourage diversity.

\newcommand{\courtinstruction}{\textbf{Task instruction:} Fill the California Small Claims Court form at \texttt{/root/sc100-blank.pdf} based on the case description below, and save the filled one to \texttt{/root/sc100-filled.pdf}. Only fill in the necessary fields and leave the court-filled, optional fields or fields not mentioned in the case description below empty. Use this date format: xxxx-xx-xx. Case Description: I am Joyce He. It's my first time suing by small claims. I live in 655 S Fair Oaks Ave, Sunnyvale, CA 94086, my phone \# is 4125886066, email: he1998@gmail.com. I want to sue Zhi Chen in 299 W Washington Ave, Sunnyvale, CA 94086. His phone \#: 5125658878. He failed to return my security deposit of amount \$1500 based on the signed roommate sublease contract after moving out. This situation happened from 2025-09-30 until 2026-01-19. I have asked him to return the money multiple times via text but he's not responding. The amount is listed on the signed roommate sublease contract. We both live in Sunnyvale, so I am filing where defendant lives. Please file it with date: January 19, 2026.}

\begin{table}[t]
  \centering
  \caption{Example query generation outputs for the \texttt{court-form-filling} task.
  The task instruction is shown at the top, followed by the generated queries for $M = 1$ (single-query prompt) and $M = 5$ (multi-query prompt).}
  \label{tab:query_examples}
  \input{tables/7_query_examples}
\end{table}

\subsection{Statistical Methodology}
\label{sec:statistical_methodology}

All confidence intervals are 95\% bootstrap percentile intervals with 10{,}000 resamples over tasks (the primary unit of analysis), using a fixed random seed for reproducibility.
For each task, the per-task Pass@$k$ is computed from the 3 runs via the unbiased combinatorial estimator \citep{chen2021evaluating}. We then bootstrap the mean across tasks to obtain CIs that reflect task-level variance, the dominant source of uncertainty with small benchmarks.

Pairwise significance is assessed via paired bootstrap tests: for each pair of conditions, per-task differences in Pass@$k$ are resampled, yielding a two-sided $p$-value as the fraction of bootstrap samples whose mean difference crosses zero.
To correct for multiple comparisons, we apply Holm-Bonferroni correction within each benchmark (3 comparisons on SkillsBench, 3 on Terminal-Bench), controlling the family-wise error rate.
Effect sizes are reported as Cohen's $h$ for proportions, with standard thresholds ($|h| < 0.20$: negligible; $< 0.50$: small; $< 0.80$: medium).

For retrieval metrics (Experiment 3), CIs are computed analogously by bootstrapping per-task recall@$k$, precision@$k$, and MRR values across the 87 SkillsBench evaluation tasks.

\subsection{Retriever Details}
\label{sec:retriever_details}

We compared several bi-encoder retriever models to determine which produces the best dense representations for skill retrieval.
We also evaluated whether embedding only the skill description or the full skill content (description + SKILL.md body) leads to better retrieval.
All models were evaluated on the SkillsBench oracle skill pairs with $k_1 = 1000$.
\Cref{tab:retriever_comparison} reports the results.
The models evaluated are:

\textbf{BAAI/bge-base-en-v1.5} \citep{xiao2024c} is a general-purpose text embedding model from the BAAI General Embedding family with 110M parameters and a maximum sequence length of 512 tokens.
It produces 768-dimensional embeddings and is trained on large-scale paired data with contrastive learning.
If the input content exceeds 512 tokens, it is truncated to fit the context length.

\textbf{BAAI/bge-m3} \citep{chen2024m3} is a multilingual, multi-functionality embedding model with 568M parameters and a maximum sequence length of 8192 tokens.
It supports dense, sparse, and multi-vector retrieval within a single model, and is trained via self-knowledge distillation.

\textbf{intfloat/e5-base-v2} \citep{wang2022text} is a text embedding model with 110M parameters and a maximum sequence length of 512 tokens.
It produces 768-dimensional embeddings and is trained with weakly-supervised contrastive pre-training on large-scale text pairs curated from the web.
If the input content exceeds 512 tokens, it is truncated to fit the context length.

\textbf{BM25} \citep{robertson2009probabilistic} is a sparse lexical retrieval method based on term frequency and inverse document frequency.

BGE-base achieves the best performance across nearly all metrics while using fewer parameters (110M) and lower latency than larger alternatives.
Embedding the description alone consistently outperforms embedding the full content for both BGE-base and BGE-M3, likely because skill descriptions are brief summaries optimized for semantic matching, whereas full content introduces noise from code examples and formatting.
BM25 performs worst, suggesting that lexical matching is poorly suited to the semantic gap between task queries and skill descriptions.
Based on these results, we select BGE-base with description-only embedding as the retriever for our pipeline.

\paragraph{Query Configuration.}
Using BGE-base, we further investigate how the number of generated queries ($M$) and the multi-query aggregation strategy affect retrieval performance.
We evaluate two aggregation methods: \textbf{Reciprocal Rank Fusion (RRF)}, which merges per-query ranked lists using reciprocal rank scores, and \textbf{Union}, which takes the union of the top-$k$ results from each query and re-ranks by the original cosine similarity.
For Union aggregation, we also vary the per-query retrieval depth $k$ (denoted $tk$).
All experiments use the multi-query prompt (v2) and BGE-base with description-only embedding.
\Cref{tab:retriever_query_config} reports the results.

With RRF aggregation, a single query ($M = 1$) outperforms multi-query configurations at top-heavy metrics (R@1 = 0.209, MRR = 0.534), while multi-query variants ($M = 3, 5$) achieve higher recall at deeper cutoffs (R@1000 = 0.884 and 0.891, respectively).
This suggests that RRF's rank-based fusion dilutes the signal of the best individual query at the top of the list, but broadens coverage at depth.

With Union aggregation, increasing $M$ from 3 to 5 consistently improves top-heavy metrics (R@1 from 0.165 to 0.174, MRR from 0.464 to 0.487) while maintaining comparable deep recall.
Varying the per-query depth $tk$ has little effect on metrics up to R@100 but affects recall at deeper cutoffs. $tk = 200$ with $M = 5$ achieves the best R@1000 (0.905).
Since the retriever is the first stage of the pipeline and its primary objective is to maximize recall, ensuring that relevant skills are not lost before the downstream reranking stages can refine the candidate set, we select Union aggregation with $M = 5$ and $tk = 200$, the configuration that achieves the highest R@1000 (0.905).

\begin{table}[t]
  \centering
  \caption{Retriever model comparison on SkillsBench oracle skill retrieval.
  Models suffixed with ``+ content'' embed the full skill content instead of the description only.
  Bold indicates best per column.
  All metrics are averaged across tasks.}
  \label{tab:retriever_comparison}
  \input{tables/8_retriever_comparison}
\end{table}

\begin{table}[t]
  \centering
  \caption{Retriever query configuration comparison (BGE-base, description-only).
  $M$ denotes the number of generated queries. $tk$ denotes the per-query retrieval depth.
  Bold indicates best per column within each aggregation group.
  All metrics are averaged across tasks.}
  \label{tab:retriever_query_config}
  \input{tables/9_retriever_query_config}
\end{table}

\paragraph{Modern retriever (recall ceiling).}
We compare the deployed \texttt{bge-base-en-v1.5} bi-encoder (110M) against \texttt{Octen-Embedding-8B} (8B), a top-ranked open-weight retriever on the RTEB and MTEB leaderboards, holding the generated query and the rest of the pipeline fixed (\Cref{tab:octen_ceiling}).
The 8B model raises recall most at R@100 (0.670 $\to$ 0.817) and at the R@1000 ceiling the cascade inherits (0.905 $\to$ 0.941).
Because embeddings are precomputed offline, the added cost is one-time reindexing and a larger index (4096-dim vs.\ 768-dim); we report the 8B as a recall ceiling and leave adopting a smaller Octen checkpoint as an efficiency-preserving deployment default to future work.

\begin{table}[t]
  \centering
  \caption{Modern retriever comparison on SkillsBench oracle skill retrieval, holding the generated query and pipeline fixed (87 tasks, 229 oracle skills).}
  \label{tab:octen_ceiling}
  \input{tables/octen_ceiling}
\end{table}

\paragraph{Structure-aware indexing.}
\Cref{tab:structure_aware} reports the 3-way YAML/prose/code sectioned index (\Cref{sec:experiment_retrieval}), encoded as three independent dense indices and late-fused at the score level.
These numbers use a matched single-query setting to isolate sectioning against the desc-only bge-base baseline (R@1000 0.843), so they are not directly comparable to the deployed multi-query retriever (R@1000 0.905).
Strategy A (3-way bge-base) improves recall at every cutoff, most at R@100, with the largest deep-recall gain from reciprocal-rank fusion (R@1000 0.843 $\to$ 0.925).
The code-encoder controls (C1/C2, and Strategy B, which uses bge-code-v1 on the code section) underperform the general-purpose bge-base, because skill retrieval here is driven by natural-language descriptions rather than code.

\begin{table}[t]
  \centering
  \caption{Structure-aware retrieval on SkillsBench (matched single-query setting).
  C1/C2 and Strategy B are code-encoder controls (bge-code-v1); Strategy A splits YAML/prose/code across three bge-base indices fused by RRF, max, mean, or sum-norm.
  Bold indicates best per column.}
  \label{tab:structure_aware}
  \input{tables/structure_aware}
\end{table}

\paragraph{In-/out-of-query oracle skills.}
To locate the source of the recall gap, we classify each oracle skill as \emph{in-query} (its description shares a content word with the generated rerank query) or \emph{out-of-query} (\Cref{tab:oracle_query_gap}).
The out-of-query share climbs roughly $10\times$ from the top-5 bucket (4\%) to the worst-ranked bucket (48\%); within the worst-ranked bucket, the 23 oracle skills entirely missing from the top-1000 are 39\% out-of-query.
The oracle skills we miss are thus largely ones the task description never names (e.g.\ \texttt{hex-grid-spatial} for a Civilization~6 adjacency task), so the residual gap is partly query generation and motivates artifact-aware query generation (\Cref{sec:discussion}).

\begin{table}[t]
  \centering
  \caption{Oracle skills bucketed by rerank rank, with the fraction that are out-of-query (description shares no content word with the generated query).
  The three buckets partition all 229 oracle skills.}
  \label{tab:oracle_query_gap}
  \input{tables/oracle_query_gap}
\end{table}

\paragraph{Stage-1 candidate count ($k_1$).}
We sweep the stage-1 cutoff $k_1 \in \{100, 250, 500, 1000, 2000, 4000\}$ (\Cref{tab:k1_sweep}).
Because the rerankers can only reorder and trim what stage~1 returns, $k_1$ sets the recall ceiling the rest of the cascade inherits.
Recall rises steeply then saturates (0.670 at 100 to 0.892 at 1000, with only ${\sim}4$ more points by 2000 and flat thereafter), and a Kneedle elbow analysis places the knee at $k_1{=}1000$.
Latency, not recall, is what makes 1000 the right operating point: of the ${\sim}35$s end-to-end (\Cref{tab:latency}), only the dense retriever (3.3s) and shallow cross-encoder (1.6s) scale with $k_1$, while the deep reranker (25.6s) runs on a fixed top-100 and the selector on its top-10.
Lowering $k_1$ therefore trims at most the 1.6s shallow stage ($<$5\% of the total) while giving up as much as 22 points of recall, so we size stage~1 for recall.
(To reach 2000 and 4000 we reran retrieval with a deeper candidate pool, so recall@1000 reads 0.892 here versus 0.905 in the deployed configuration.)

\begin{table}[t]
  \centering
  \caption{Stage-1 candidate-count ($k_1$) sweep on SkillsBench retrieval (87 tasks, 229 oracle skills).
  Recall@$k_1$ is the cascade ceiling the downstream rerankers inherit.}
  \label{tab:k1_sweep}
  \input{tables/k1_sweep}
\end{table}

\subsection{Reranker Details}
\label{sec:reranker_details}

We evaluate cross-encoder reranker models and their configurations to select the best setup for the shallow and deep reranking stages.
We first compare two cross-encoder models on the full 1000-candidate retriever output to determine model selection, then investigate the effect of input chunk size on the deep reranker operating on the 100-candidate shallow reranker output.
The two cross-encoder models compared are:

\textbf{cross-encoder/ms-marco-MiniLM-L-6-v2} \citep{reimers2019sentence} is a lightweight cross-encoder with 6 Transformer layers and a maximum sequence length of 512 tokens, trained on the MS MARCO passage ranking dataset for fast inference.

\textbf{BAAI/bge-reranker-v2-m3} \citep{chen2024m3} is a multilingual cross-encoder reranker with a maximum sequence length of 8192 tokens, trained with multi-stage knowledge distillation.

\paragraph{Model Comparison.}
We compare both models on the 1000-candidate retriever output with skill content truncated to 512 characters.
We vary the number of queries ($M \in \{1, 3, 5\}$) and, for multi-query configurations, the score aggregation strategy: \textbf{mean} (average relevance score across queries), \textbf{max} (maximum score across queries), and \textbf{RRF} (reciprocal rank fusion).
Single-query ($M = 1$) configurations require no aggregation.
\Cref{tab:reranker_comparison} reports the results.

BGE-reranker-v2-m3 with a single query ($M = 1$) achieves the best performance across nearly all metrics, with an MRR of 0.649, R@1 of 0.276, and R@50 of 0.720.
However, BGE-reranker-v2-m3 requires ${\sim}$12s per task to score 1000 candidates, approximately $10.6\times$ the inference time of ms-marco-MiniLM (${\sim}$1.1s), making it impractical for the shallow reranking stage, which must score 1000 candidates in near real-time.
The second-best configuration, ms-marco-MiniLM with $M = 1$ (underlined in \Cref{tab:reranker_comparison}), achieves competitive performance (MRR = 0.587, R@1 = 0.259) at a fraction of the cost.
For both models, single-query scoring consistently outperforms multi-query configurations on top-heavy metrics, echoing the finding from the retriever stage (\Cref{sec:retriever_details}).
Among multi-query aggregation strategies, mean performs best, followed by max and RRF.
We therefore select ms-marco-MiniLM-L-6-v2 with $M = 1$ for the shallow reranker (1000 $\rightarrow$ 100 candidates) and reserve BGE-reranker-v2-m3 for the deep reranker (100 $\rightarrow$ 10 candidates), where the smaller candidate set makes its higher computational cost acceptable.

\begin{table}[t]
  \centering
  \caption{Reranker model comparison on SkillsBench oracle skill retrieval.
  Both models are evaluated on the 1000-candidate retriever output with skill content truncated to 512 characters.
  $M$ denotes the number of generated queries. Aggregation strategy applies only when $M > 1$.
  Bold indicates best per column. Underline indicates the selected configuration for the shallow reranker.
  All metrics are averaged across tasks.}
  \label{tab:reranker_comparison}
  \input{tables/10_reranker_comparison}
\end{table}

\paragraph{Stronger and longer-context rerankers.}
On the same 1000-candidate retriever set (already at R@1000 0.905) and the same generated query, we evaluate stronger and longer-context cross-encoders (\Cref{tab:reranker_modern}).
R@100 is essentially flat across rerankers (0.776--0.786), so R@100 is retriever-bound and the reranker is not the recall bottleneck; a stronger cross-encoder mainly helps top-heavy ranking (bge-reranker-v2-m3 raises MRR from 0.587 to 0.649 and P@1 from 0.494 to 0.563 over ms-marco).
Qwen3-Reranker-0.6B does not beat ms-marco at 512 chars and only matches bge's P@1 at 1024 chars, at higher latency and still without lifting R@100.
The shallow cross-encoder is thus a latency/precision lever with little effect on recall, consistent with prior findings that a cross-encoder's recall contribution is bounded once the candidate set is fixed.

\begin{table}[t]
  \centering
  \caption{Stronger and longer-context rerankers on the same 1000-candidate retriever set and generated query.
  R@1000 (0.905) is fixed by the retriever candidate set. Bold indicates best per column.}
  \label{tab:reranker_modern}
  \input{tables/reranker_modern}
\end{table}

\paragraph{Deep Reranker Configuration.}
Using BGE-reranker-v2-m3 on the 100-candidate output of the shallow reranker, we investigate the effect of input chunk size, the character truncation length applied to skill content, on reranking quality.
We evaluate chunk sizes of 1024, 2048, 4096, and 8192 characters, along with query count ($M$) and aggregation strategy.
\Cref{tab:deep_reranker_config} reports results for all configurations.

\begin{figure}[t]
  \centering
  \includegraphics[width=\columnwidth]{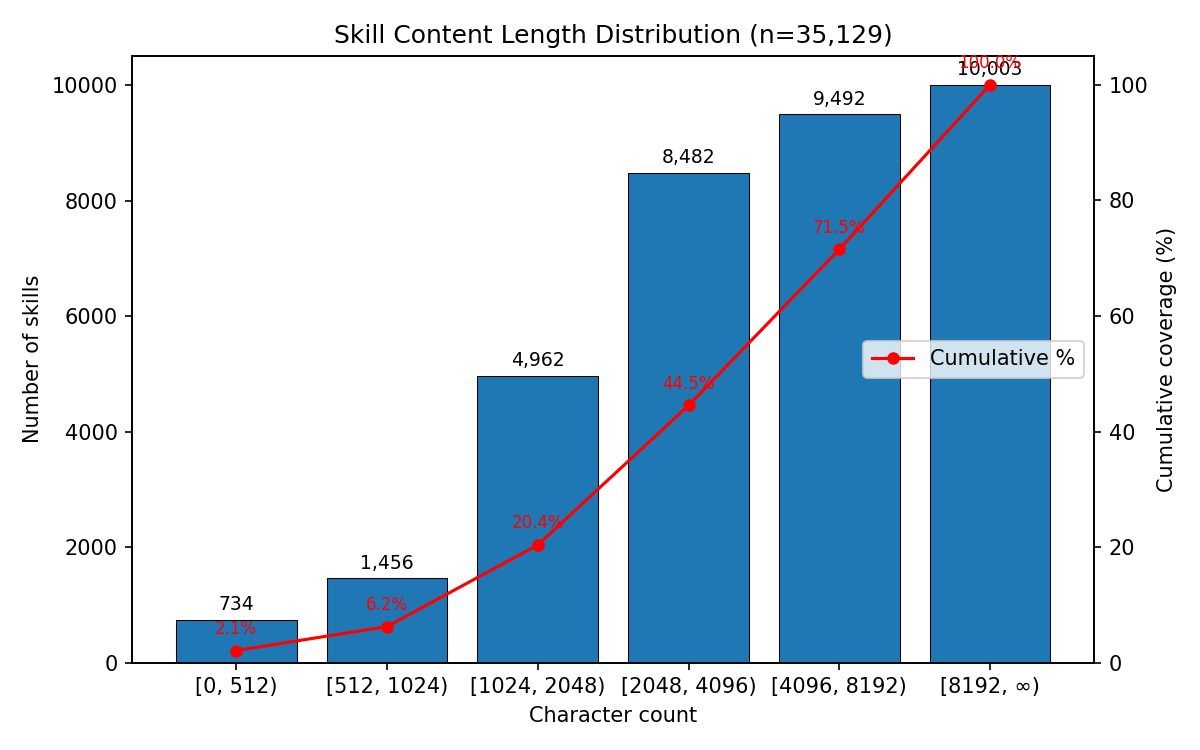}
  \caption{Distribution of skill content length (in characters) across the 35,129 indexed skills in the library.
  The red line shows cumulative coverage: a 4096-character window preserves 44.5\% of skills in full.}

  \label{fig:skill_content_dist}
\end{figure}

Among single-query configurations, MRR decreases monotonically as chunk size increases (0.653 at 1024 to 0.620 at 8192), while R@5 and R@10 peak at 4096 characters (0.499 and 0.595, respectively).
Skill content is long relative to these windows: as shown in \Cref{fig:skill_content_dist}, a 4096-character truncation preserves only 44.5\% of skills in full, and even 8192 characters reaches just 71.5\%.
Coverage alone therefore does not explain the optimum.
Shorter truncations (1024 and 2048 characters, preserving 6.2\% and 20.4\% of skills) discard informative content and hurt R@5 and R@10, whereas extending to 8192 characters adds substantial coverage yet degrades every metric, indicating that the extra tail content dilutes the relevance signal more than it informs it.
A 4096-character window balances the two: it captures the informative head of most skills without admitting the noisy remainder.

Multi-query configurations underperform single-query across all chunk sizes and both reranker models.
Unlike the retrieval stage, where multiple queries broaden coverage over 35K skills, the reranking stage operates on a small, already-filtered candidate set where a single focused query is sufficient to discriminate among candidates and additional queries likely introduce more noise than signal.

Based on these results, we select BGE-reranker-v2-m3 with 4096-character truncation and a single query ($M = 1$) as the deep reranker configuration, as it achieves the best R@5 (0.499) and R@10 (0.595), the metrics most relevant for the deep reranker's output size of $k_3 = 10$, while maintaining strong MRR (0.634).

\begin{table}[t]
  \centering
  \caption{Deep reranker configuration comparison (BGE-reranker-v2-m3) on the 100-candidate shallow reranker output.
  Chunk size denotes the character truncation length applied to skill content.
  Bold indicates best per column.
  All metrics are averaged across tasks.}
  \label{tab:deep_reranker_config}
  \input{tables/11_deep_reranker_config}
\end{table}

\subsection{Selector Details}
\label{sec:selector_details}

The selector is a two-step LLM-based filter that operates on the top-$k$ candidates (default $k{=}10$) from the deep reranker.
It determines which retrieved skills, if any, should be injected into the agent's workspace.

\paragraph{Step 1: Relevancy Filter.}
An LLM (gpt-4o-mini) receives the task instruction and each candidate skill's full \texttt{SKILL.md} content.
It makes a binary judgment: is this skill topically relevant to the task?
The LLM outputs a JSON list of selected candidate indices.
Skills that are off-topic or unrelated are removed.
The system instruction (\Cref{box:selector_system_prompt}) and user instruction (\Cref{box:selector_user_prompt}) structure this as a classification task.

\paragraph{Step 2: Specificity Filter.}
Skills that pass relevancy are evaluated for actionable specificity.
A second LLM call (same model) assesses whether the skill provides domain-specific knowledge that would save the agent meaningful exploration time, as opposed to generic methodology, checklists, or information the agent can derive from its training data or the task environment.
The specificity instruction (\Cref{box:specificity_prompt}) defines the INJECT/REJECT criteria.

\paragraph{No-selector ablation.}
To test whether the selector is necessary, we compare full SkillFlow against injecting the deep reranker's top-10 skills directly, with no selector (\Cref{tab:selector_ablation}).
Downstream Pass@1 is statistically indistinguishable (16.4\% vs.\ 15.9\%, paired bootstrap $p = 0.91$) and cost is unchanged (\$0.025 vs.\ \$0.026, $p = 0.50$), while both remain well above the no-skill baseline (9.2\%).
The selector's role is therefore context compaction, shrinking the injected set from 10 to a mean of 2.8 skills (\Cref{tab:adoption}), rather than accuracy, so we treat it as an optional gate (\Cref{sec:method}, \Cref{sec:experiment_benchmark}).

\begin{table}[t]
  \centering
  \caption{No-selector ablation on SkillsBench (Codex, GPT-5-mini).
  Deep-reranker-only injects the deep reranker's top-10 skills directly; full SkillFlow adds the LLM selector. Paired bootstrap $p$-values vs.\ full SkillFlow are reported in the text.}
  \label{tab:selector_ablation}
  \input{tables/20_selector_ablation}
\end{table}

\begin{figure}[!p]
\begin{tcolorbox}[title=Selector System Prompt (Relevancy), colback=gray!5, colframe=gray!50, fonttitle=\bfseries\small, fontupper=\scriptsize\ttfamily\raggedright]
You are a skill utility judge for an AI coding agent. The agent will receive at most ONE candidate "skill" (an instruction document) to help it solve a task. Your job is to decide which single skill, if any, is worth injecting.\newline

IMPORTANT: Injecting skills has a COST: approach bias and distraction. The agent solves most tasks BETTER without skills. Only inject a skill if you are confident it will help more than it hurts.\newline

\#\# Two-phase evaluation\newline

\#\#\# Phase 1: Does the agent need external help?\newline

Read the task description and ask:\newline

"Does this task require specific procedural knowledge that can only come from hands-on experience with the exact tools/environment, or can the agent solve it from first principles?"\newline

The agent is an expert-level AI with deep knowledge of programming, algorithms, frameworks, and well-documented tools.\newline

The agent does NOT need help when:\newline
- The task has a clear logical or algorithmic solution, even if the domain sounds specialized.\newline
- The task involves well-documented tools or formats.\newline
- The task description specifies what to do clearly enough that reasoning is sufficient.\newline

The agent DOES need help when:\newline
- The task requires environment-specific procedural knowledge: exact command sequences, build system workarounds, boot sequences, or binary format internals.\newline
- The task involves tool interactions with non-obvious failure modes: undocumented behavior, version-specific bugs, or subtle compatibility issues.\newline

If the agent likely does NOT need help → select nothing.\newline

\#\#\# Phase 2: Pick the single best skill (only if the agent needs help)\newline

From the candidates, select at most one: the single skill that most precisely matches the task's core difficulty.\newline

KEEP a skill only if:\newline
1. It addresses the exact difficulty identified in Phase 1.\newline
2. It provides concrete, non-obvious procedural details (exact commands, specific parameter values, workarounds).\newline
3. It is precisely aligned with the task, not a generic guide for a related topic.\newline

REJECT a skill if:\newline
1. It restates knowledge the agent already has.\newline
2. It provides general methodology rather than specific procedures.\newline
3. It could mislead the agent if task details differ from the skill's assumptions.\newline

\#\# Output\newline

Return a JSON object: \{"selected": [N]\} where N is the single best candidate number, or \{"selected": []\} if no skill is worth injecting.
\end{tcolorbox}
\caption{System prompt for the selector relevancy filter (Step~1).
The LLM classifies each candidate skill as relevant or irrelevant to the task.}

\label{box:selector_system_prompt}
\end{figure}

\begin{figure}[!p]
\begin{tcolorbox}[title=Selector User Prompt (Shared), colback=gray!5, colframe=gray!50, fonttitle=\bfseries\small, fontupper=\scriptsize\ttfamily\raggedright]
\#\# Task\newline
\{\{ query \}\}\newline

\#\# Candidate Skills\newline
\{\% for candidate in candidates \%\}\newline
\#\#\# Candidate \{\{ loop.index \}\}\newline
\{\{ candidate.content or candidate.description \}\}\newline
\{\% endfor \%\}
\end{tcolorbox}
\caption{User prompt template shared by both selector steps.
The task instruction and candidate skill contents are populated via Jinja2 templating.}

\label{box:selector_user_prompt}
\end{figure}

\begin{figure}[!p]
\begin{tcolorbox}[title=Selector Specificity Prompt (Step~2), colback=gray!5, colframe=gray!50, fonttitle=\bfseries\small, fontupper=\scriptsize\ttfamily\raggedright]
You are a second-pass filter for AI agent skill injection. The candidate skill below has already been judged as topically relevant. Your job is to determine whether injecting it would save the agent meaningful time.\newline

\#\# Context\newline

The agent is an expert-level AI coding assistant. It has broad knowledge of programming languages, frameworks, and common tools. It can also explore the task environment: reading files, running commands, and checking documentation.\newline

However, the agent works under a time budget. Skills are valuable when they reduce the exploration and trial-and-error the agent would otherwise need.\newline

\#\# Decision criteria\newline

\#\#\# INJECT: the skill provides domain-specific knowledge that would take the agent significant effort to assemble on its own:\newline

- Specific API usage patterns, method signatures, or library workflows for specialized tools\newline
- Domain-specific configuration, parameters, or data formats\newline
- Multi-step procedures that require combining knowledge from multiple sources\newline
- Tool-specific idioms or integration patterns not obvious from --help alone\newline
- Concrete examples, code snippets, or templates for niche tasks\newline

\#\#\# REJECT: the skill adds no time savings:\newline

- Pure methodology or approach guidance without concrete implementation details\newline
- Checklists or best-practice lists without actionable specifics\newline
- Content that merely restates or reorganizes the task description\newline
- Vague overviews that lack concrete code, commands, or configurations\newline

When in doubt, INJECT. A marginally useful skill is better than a missed one.\newline

\#\# Output\newline

Return \{"selected": [N]\} if the skill passes, or \{"selected": []\} if it does not.
\end{tcolorbox}
\caption{Specificity filter prompt (Step~2).
Evaluates whether a relevant skill provides actionable, domain-specific knowledge (INJECT) or only generic guidance (REJECT).}

\label{box:specificity_prompt}
\end{figure}

\subsection{Vercel Baseline Details}
\label{sec:vercel_details}

We compare against skills.sh, Vercel's public skill search API, as a commercial baseline.
To ensure a fair comparison, we reuse the same LLM-generated queries from SkillFlow's cache.
For each task, the cached query is sent to the skills.sh REST API (\texttt{GET https://skills.sh/api/search?q=<query>}), and the top-1 non-duplicate result is selected.
Each matched skill is then downloaded via Vercel's official CLI (\texttt{npx skills add <source> -s <skillId> -y {-}{-}copy -a claude-code}), which fetches the full skill folder from the source GitHub repository.
Downloaded skills are evaluated using the same Harbor-based harness, agent model, and concurrency settings as all other conditions, with 1:1 task-to-skill matching by task name.

\subsection{Task Exclusion Criteria}
\label{sec:task_exclusion}

We exclude tasks from both benchmarks when they cannot be evaluated reliably in our sandboxed Docker environment.
Exclusion decisions were made during pilot runs based on three criteria: (1)~\textbf{infrastructure} errors, where the Daytona cloud sandbox failed to provision the task environment or the Docker build failed (e.g., OOM during model loading, intermittent downloads, hardcoded host paths), with Docker build failures documented on the SkillsBench website; (2)~\textbf{reproducibility} issues, where the task's oracle verifier produces intermittent failures due to environment-sensitive tests, flaky builds, or external API dependencies, also documented on the SkillsBench website; and (3)~\textbf{resource constraints}, where the task requires large corpus downloads or model training exceeding the sandbox budget.
\Cref{tab:excluded_tasks} lists all excluded tasks.

\begin{table}[t]
  \centering
  \caption{Excluded benchmark tasks.
  Of 87 SkillsBench tasks, 22 are excluded (65 retained).
  Of 89 Terminal-Bench tasks, 1 is excluded (88 retained).}
  \label{tab:excluded_tasks}
  \input{tables/12_excluded_tasks}
\end{table}

\subsection{Qualitative Case Studies}
\label{sec:case_studies}

\paragraph{Case Study 1: Invoice Fraud Detection.}
\label{sec:case_study_1}

\paragraph{Task.}
Injected skills: \texttt{fuzzy-match}, \texttt{pdf}, \texttt{xlsx}.
Outcome: 2/2 tests passed.

\begin{table}[t]
  \centering
  \caption{Pipeline progression for \texttt{invoice-fraud-detection}.}
  \label{tab:case_study_1}
  \input{tables/13_case_study_1}
\end{table}

\paragraph{Agent behavior.}
The dense retriever returns generic PDF extraction skills, none directly relevant to fraud detection.
The cross-encoder reranker surfaces \texttt{fuzzy-matching} (a community skill) to rank~1 with a score of 3.74, pushing it ahead of 689 candidates.
The deep reranker confirms this ranking (0.911) and the LLM selector marks it as both relevant and specific.
The agent used all three injected skills (\texttt{fuzzy-match}, \texttt{pdf}, \texttt{xlsx}) and passed all tests.
This case illustrates how reranking transforms an uninformative Stage~1 ranking into actionable skill selection.

\paragraph{Case Study 2: Court Form Filling.}
\label{sec:case_study_2}

\paragraph{Task.} Injected skills: \texttt{pdf}. Outcome: 0/5 tests passed.

\begin{table}[t]
  \centering
  \caption{Pipeline progression for \texttt{court-form-filling}.}
  \label{tab:case_study_2}
  \input{tables/14_case_study_2}
\end{table}

\paragraph{Agent behavior.}
All pipeline stages surface PDF-related skills: \texttt{python-pdf} is the top candidate by Stage~3 (score 0.706) and the only skill the selector marks as relevant.
The agent used the injected \texttt{pdf} skill, which provides general PDF manipulation guidance (e.g., PyPDF2 API usage).
However, the task requires filling specific fields in a California Small Claims Court form (SC-100), a domain-specific mapping that no generic PDF skill can provide.
All 5 tests failed.
This case illustrates the ``skill quality is the bottleneck'' finding: the pipeline correctly identifies the technology domain (PDF manipulation) but the library lacks a sufficiently specific skill for legal form filling.

\paragraph{Case Study 3: Gravitational Wave Detection.}
\label{sec:case_study_3}

\paragraph{Task.}
Injected skills: \texttt{conditioning}, \texttt{matched-filtering}.
Outcome: 6/9 tests passed.

\begin{table}[t]
  \centering
  \caption{Pipeline progression for \texttt{gravitational-wave-detection}.}
  \label{tab:case_study_3}
  \input{tables/15_case_study_3}
\end{table}

\paragraph{Agent behavior.}
Retrieval is near-perfect: the oracle skill \texttt{matched-filtering} ranks first at every stage (scores: 0.844 $\rightarrow$ 7.51 $\rightarrow$ 0.999), and \texttt{conditioning} ranks second throughout.
However, the LLM selector marks \emph{all} 10 candidates as not relevant (relevancy=0), including both oracle skills.
Despite this, the agent used both injected skills and passed 6 of 9 tests, failing on the most complex signal processing subtasks.
This case highlights two issues: (1) the LLM selector can be overly conservative with domain-specific terminology, and (2) even with perfect retrieval and skill adoption, execution complexity limits performance.

\subsection{Retrieved-Skill Usage}
\label{sec:skill_usage}

We analyze how agents use retrieved skills across the SkillsBench runs, separating oracle from non-oracle retrievals (\Cref{tab:skill_usage}).
Non-oracle skills make up 43\% of the injected set but are opened only 17\% of the time, versus 42\% for oracle skills, about half as often.
Relative to the no-skill baseline, retrieved skills improve 16 of 64 tasks and regress 4 (per-task mean Pass over three runs, paired bootstrap $p = 0.02$), and the improvement concentrates in tasks where retrieval recovered at least one oracle skill.
In the four tasks where only non-oracle skills were retrieved, the agent loads them but solves none, and the clearest regression (\texttt{jax-computing}) is one where the agent fixates on a non-oracle skill and never opens the oracle skill.
Non-oracle skills are thus mostly not useful as-is but rarely degrade performance, which motivates skill adaptation as future work.

\begin{table}[t]
  \centering
  \caption{Retrieved-skill usage on SkillsBench (per-task mean Pass over three runs).
  Rows break tasks down by what SkillFlow injected; Helped/Hurt are per-task counts relative to the no-skill baseline.
  Oracle vs.\ non-oracle load rates are reported in the text.}
  \label{tab:skill_usage}
  \input{tables/18_skill_usage}
\end{table}

\subsection{Oracle Ceiling Analysis}
\label{sec:oracle_ceiling}

To understand why even perfect retrieval yields only 19.5\% Pass@1 on SkillsBench, we examine three converging factors evident in our results.
First, \emph{agent non-adoption}: despite oracle skills being available for every task, agents use at least one skill in only 69.2\% of tasks, meaning 30.8\% of tasks receive no skill benefit at all because the agent judges the provided skills as unnecessary.
This likely reflects a mismatch between the skills' instructional style (general-purpose reference documents) and what the agent recognizes as actionable for a specific task instance.
Second, \emph{task inherent difficulty}: the no-skill baseline itself stands at only 9.2\% Pass@1, confirming that SkillsBench tasks are challenging for the agent model regardless of skill availability.
Many tasks involve multi-step workflows (e.g., filling legal PDF forms, configuring complex build systems) where a single misstep in tool use or reasoning cascades into failure, and a reference document alone cannot compensate for these execution errors.
Third, \emph{low consistency across runs}: the oracle condition achieves Pass@3 of 35.4\% but Pass\textasciicircum3 of only 6.2\%, indicating that most oracle-aided successes are non-deterministic: the agent solves the task in some runs but not others.
This high variance suggests that even when skills do help, the benefit is fragile and depends on the specific reasoning trace the agent follows.

Together, these factors delineate a ceiling imposed not by retrieval quality but by the interaction between skill format, agent reasoning, and task complexity.
Closing this gap will require advances on multiple fronts: skills that provide more directly executable artifacts rather than reference prose, agents that can more reliably incorporate instructional content into their planning, and evaluation protocols that account for the stochastic nature of agent execution.

\end{document}

%% file: tables/1_results.tex
\resizebox{\columnwidth}{!}{%
\begin{tabular}{llccccc}
  \toprule
  \textbf{Model} & \textbf{Condition} & \textbf{Pass@1} & \textbf{Pass@3} & \textbf{Pass\textasciicircum3} & \textbf{Steps/Task} & \textbf{Cost/Task} \\
  \midrule
  \multicolumn{7}{l}{\textit{SkillsBench}} \\
  \midrule
  GPT-5-mini       & \textit{Oracle}  & \textit{19.5}$^{*}${\scriptsize~[12.3, 27.2]} & \textit{35.4}{\scriptsize~[24.6, 47.7]} & \textit{6.2}{\scriptsize~[1.5, 12.3]} & \textit{20.1} & \textit{\$0.028} \\[2pt]
                   & No Skills        & 9.2{\scriptsize~[4.1, 14.9]} & 16.9{\scriptsize~[7.7, 26.2]} & 0.0{\scriptsize~[0.0, 0.0]} & \textbf{18.7} & \$0.026 \\
                   & Vercel           & 9.7{\scriptsize~[4.6, 15.4]} & 20.0{\scriptsize~[10.8, 30.8]} & 1.5{\scriptsize~[0.0, 4.6]} & 20.9 & \$0.029 \\
                   & SkillFlow (Ours) & \textbf{16.4}$^{*}${\scriptsize~[9.7, 23.6]} & \textbf{29.2}{\scriptsize~[18.5, 40.0]} & \textbf{4.6}{\scriptsize~[0.0, 10.8]} & 18.8 & \textbf{\$0.025} \\
  \cmidrule(lr){1-7}
  Claude Haiku 4.5 & \textit{Oracle}  & \textit{30.3}$^{**}${\scriptsize~[21.5, 39.5]} & \textit{44.6}{\scriptsize~[32.3, 56.9]} & \textit{13.8}{\scriptsize~[6.2, 23.1]} & \textit{55.4} & \textit{\$0.428} \\[2pt]
                   & No Skills        & 6.7{\scriptsize~[2.6, 11.8]} & 12.3{\scriptsize~[4.6, 21.5]} & 0.0{\scriptsize~[0.0, 0.0]} & \textbf{50.8} & \$0.418 \\
                   & SkillFlow (Ours) & \textbf{17.4}$^{**}${\scriptsize~[10.3, 25.1]} & \textbf{29.2}{\scriptsize~[18.5, 40.0]} & \textbf{4.6}{\scriptsize~[0.0, 10.8]} & 51.4 & \textbf{\$0.404} \\
  \midrule
  \multicolumn{7}{l}{\textit{Terminal-Bench}} \\
  \midrule
  GPT-5-mini       & No Skills        & 34.8{\scriptsize~[26.1, 43.6]} & 46.6{\scriptsize~[36.4, 56.8]} & 20.5{\scriptsize~[12.5, 29.5]} & \textbf{21.1} & \textbf{\$0.030} \\
                   & Vercel           & 32.6{\scriptsize~[24.2, 41.3]} & 45.5{\scriptsize~[35.2, 55.7]} & \textbf{21.6}{\scriptsize~[13.6, 30.7]} & 21.2 & \$0.031 \\
                   & SkillFlow (Ours) & 31.8{\scriptsize~[23.9, 40.2]} & 46.6{\scriptsize~[36.4, 56.8]} & 17.0{\scriptsize~[9.1, 25.0]} & 23.0 & \$0.034 \\
                   & SkillFlow-specific (Ours) & \textbf{34.9}{\scriptsize~[26.7, 43.2]} & \textbf{48.0}{\scriptsize~[37.8, 58.0]} & 21.0{\scriptsize~[13.4, 29.0]} & 24.2 & \$0.035 \\
  \bottomrule
\end{tabular}%
}

%% file: tables/2_adoption.tex
\resizebox{\columnwidth}{!}{%
\begin{tabular}{rcccc}
  \toprule
  \textbf{Condition} & \textbf{Mean Skills Retrieved/Task} & \textbf{Tasks Retrieved (\%)} & \textbf{Oracle Skills Retrieved (\%)} & \textbf{Tasks Used (\%)} \\
  \midrule
  \multicolumn{5}{l}{\textit{SkillsBench}} \\
  \midrule
  \textit{Oracle}          & \textit{2.5} & \textit{100.0{\scriptsize~[100.0, 100.0]}} & \textit{100.0{\scriptsize~[100.0, 100.0]}} & \textit{69.2{\scriptsize~[62.6, 75.9]}} \\
  Vercel                   & 0.9  & 87.5{\scriptsize~[82.8, 92.2]} & ---   & 40.6{\scriptsize~[33.9, 47.9]} \\
  SkillFlow top-1 (Ours)   & 0.9  & 91.7{\scriptsize~[87.5, 95.3]} & 37.6{\scriptsize~[32.4, 43.0]} & \textbf{68.8}$^{**}${\scriptsize~[62.5, 75.0]} \\
  SkillFlow (Ours)         & 2.8  & \textbf{92.2}{\scriptsize~[88.0, 95.8]} & \textbf{61.8}{\scriptsize~[56.5, 66.9]} & \textbf{68.8}$^{**}${\scriptsize~[62.0, 75.0]} \\
  \midrule
  \multicolumn{5}{l}{\textit{Terminal-Bench}} \\
  \midrule
  Vercel                   & 0.9  & \textbf{90.9}{\scriptsize~[87.1, 94.3]} & ---   & 22.3{\scriptsize~[17.4, 27.7]} \\
  SkillFlow top-1 (Ours)   & 0.9  & 89.8{\scriptsize~[86.0, 93.2]} & ---   & \textbf{72.7}$^{**}${\scriptsize~[67.4, 78.0]} \\
  SkillFlow (Ours)         & 1.5  & 87.5{\scriptsize~[83.3, 91.3]} & ---   & 70.1$^{**}${\scriptsize~[64.4, 75.4]} \\
  \bottomrule
\end{tabular}%
}

%% file: tables/3_retrieval_stages.tex
\resizebox{\columnwidth}{!}{%
\begin{tabular}{ccccccccccc}
  \toprule
  \textbf{Stage} & \textbf{K output} & \textbf{Metric} & \textbf{@1} & \textbf{@5} & \textbf{@10} & \textbf{@50} & \textbf{@100} & \textbf{@500} & \textbf{@1000} & \textbf{MRR} \\
  \midrule
  \multirow{2}{*}{Retriever}        & \multirow{2}{*}{1000}    & R   & 0.174{\scriptsize~[0.12, 0.24]} & 0.376{\scriptsize~[0.30, 0.46]} & 0.469{\scriptsize~[0.39, 0.55]} & 0.603{\scriptsize~[0.52, 0.69]} & 0.670{\scriptsize~[0.59, 0.75]} & 0.859{\scriptsize~[0.80, 0.91]} & 0.905{\scriptsize~[0.85, 0.95]} & \multirow{2}{*}{0.487{\scriptsize~[0.40, 0.58]}} \\
                                    &                          & P   & 0.379{\scriptsize~[0.28, 0.48]} & 0.182{\scriptsize~[0.14, 0.22]} & 0.122{\scriptsize~[0.10, 0.15]} & 0.032{\scriptsize~[0.03, 0.04]} & 0.018{\scriptsize~[0.02, 0.02]} & 0.005{\scriptsize~[0.00, 0.01]} & 0.002{\scriptsize~[0.00, 0.00]} &  \\
  \midrule
  \multirow{2}{*}{Shallow Reranker} & \multirow{2}{*}{100}     & R   & 0.259{\scriptsize~[0.19, 0.33]} & 0.460{\scriptsize~[0.38, 0.55]} & 0.520{\scriptsize~[0.43, 0.61]} & 0.687{\scriptsize~[0.61, 0.76]} & 0.776{\scriptsize~[0.71, 0.84]} & ---   & ---   & \multirow{2}{*}{0.587{\scriptsize~[0.50, 0.68]}} \\
                                    &                          & P   & 0.494{\scriptsize~[0.39, 0.60]} & 0.216{\scriptsize~[0.18, 0.26]} & 0.130{\scriptsize~[0.11, 0.16]} & 0.034{\scriptsize~[0.03, 0.04]} & 0.019{\scriptsize~[0.02, 0.02]} & ---   & ---   &  \\
  \midrule
  \multirow{2}{*}{Deep Reranker}    & \multirow{2}{*}{10}      & R   & 0.271{\scriptsize~[0.20, 0.34]} & 0.499{\scriptsize~[0.42, 0.58]} & 0.595{\scriptsize~[0.51, 0.68]} & ---   & ---   & ---   & ---   & \multirow{2}{*}{0.634{\scriptsize~[0.54, 0.72]}} \\
                                    &                          & P   & 0.540{\scriptsize~[0.44, 0.64]} & 0.237{\scriptsize~[0.20, 0.28]} & 0.147{\scriptsize~[0.12, 0.17]} & ---   & ---   & ---   & ---   &  \\
  \midrule
  \multirow{2}{*}{Selector}         & \multirow{2}{*}{$\leq$5} & R   & 0.273{\scriptsize~[0.20, 0.34]} & 0.455{\scriptsize~[0.38, 0.53]} & ---   & ---   & ---   & ---   & ---   & \multirow{2}{*}{0.639{\scriptsize~[0.55, 0.73]}} \\
                                    &                          & P   & 0.563{\scriptsize~[0.46, 0.67]} & 0.221{\scriptsize~[0.18, 0.26]} & ---   & ---   & ---   & ---   & ---   &  \\
  \bottomrule
\end{tabular}%
}

%% file: tables/4_stage_ablation.tex
\resizebox{\columnwidth}{!}{%
\begin{tabular}{lccccc}
  \toprule
  \textbf{Pipeline} & \textbf{MRR} & \textbf{R@5} & \textbf{R@10} & \textbf{P@10} & \textbf{Hit@10} \\
  \midrule
  \multicolumn{6}{l}{\textit{First-stage alternatives}} \\
  \midrule
  BM25 only                        & 0.266{\scriptsize~[0.19, 0.35]} & 0.211{\scriptsize~[0.14, 0.29]} & 0.238{\scriptsize~[0.17, 0.32]} & 0.055{\scriptsize~[0.04, 0.07]} & 0.391{\scriptsize~[0.29, 0.49]} \\
  Hybrid (Dense $+$ BM25)          & 0.522{\scriptsize~[0.43, 0.61]} & 0.411{\scriptsize~[0.33, 0.49]} & 0.480{\scriptsize~[0.40, 0.56]} & 0.114{\scriptsize~[0.09, 0.14]} & 0.713{\scriptsize~[0.61, 0.81]} \\
  Dense only                       & 0.553{\scriptsize~[0.46, 0.64]} & 0.415{\scriptsize~[0.34, 0.50]} & 0.477{\scriptsize~[0.39, 0.56]} & 0.121{\scriptsize~[0.10, 0.15]} & 0.713{\scriptsize~[0.61, 0.81]} \\
  \midrule
  \multicolumn{6}{l}{\textit{Cumulative pipeline}} \\
  \midrule
  $+$ Shallow reranker (1--2)      & 0.587{\scriptsize~[0.50, 0.68]} & 0.460{\scriptsize~[0.38, 0.55]} & 0.520{\scriptsize~[0.43, 0.61]} & 0.130{\scriptsize~[0.11, 0.16]} & 0.724{\scriptsize~[0.63, 0.82]} \\
  $+$ Deep reranker (1--3)         & 0.634{\scriptsize~[0.54, 0.72]} & \textbf{0.499}{\scriptsize~[0.42, 0.58]} & \textbf{0.595}{\scriptsize~[0.51, 0.68]} & \textbf{0.147}{\scriptsize~[0.12, 0.17]} & \textbf{0.793}{\scriptsize~[0.70, 0.87]} \\
  $+$ Selector (1--4)              & \textbf{0.639}{\scriptsize~[0.55, 0.73]} & 0.455{\scriptsize~[0.38, 0.53]} & --- & --- & --- \\
  \bottomrule
\end{tabular}%
}

%% file: tables/5_latency.tex
\begin{tabular}{lrrr}
  \toprule
  \textbf{Stage} & \textbf{Median} & \textbf{Mean} & \textbf{P95} \\
  \midrule
  Stage 1: Dense Retriever        & 3.3 & 3.4 & 4.0 \\
  Stage 2: Cross-Encoder Reranker & 1.6 & 1.6 & 2.2 \\
  Stage 3: Deep Reranker          & 25.6 & 26.4 & 33.1 \\
  Stage 4: LLM Selector           & 2.7 & 3.6 & 8.5 \\
  \midrule
  \textbf{Total}                  & \textbf{35.0} & \textbf{35.0} & \textbf{43.8} \\
  \bottomrule
\end{tabular}

%% file: tables/6_corpus_stats.tex
\begin{tabular}{lr}
  \toprule
  \textbf{Metric} & \textbf{Count} \\
  \midrule
  Skills processed              & 43,660 \\
  Excluded (repo $>$ 50\,MB)    & 7,703 \\
  Failed (deleted/inaccessible) & 91 \\
  Downloaded                    & 35,866 \\
  Indexed (valid SKILL.md)      & 35,129 \\
  \bottomrule
\end{tabular}

%% file: tables/19_functional_audit.tex
\resizebox{\columnwidth}{!}{%
\begin{tabular}{lr}
  \toprule
  \textbf{Metric} & \textbf{Value} \\
  \midrule
  Code-bearing (corpus, n=35,129) & 78.1\% \\
  Bundles executable scripts & 13.2\% \\
  Has fenced code block & 76.5\% \\
  Has any bundled file & 39.1\% \\
  \midrule
  Runnable-complete (sample, n=300) & 32.3\% {\scriptsize~[27.3\%, 37.8\%]} \\
  \quad judge tier: functional & 98 \\
  \quad judge tier: partial & 26 \\
  \quad judge tier: reference-only & 176 \\
  \bottomrule
\end{tabular}%
}

%% file: tables/7_query_examples.tex
\begin{tabular}{p{1.5cm}p{11cm}}
  \toprule
  \multicolumn{2}{p{12.5cm}}{\courtinstruction} \\
  \midrule
  $M = 1$ & California Small Claims Court form filling automation using Python PDF libraries \\
  \midrule
  $M = 5$ & 1.~California Small Claims Court form filling \newline 2.~PDF form automation with Python \newline 3.~Filling out legal documents using PyPDF2 \newline 4.~Using ReportLab for PDF generation in Python \newline 5.~Creating and editing PDF forms with pdfrw \\
  \bottomrule
\end{tabular}

%% file: tables/8_retriever_comparison.tex
\resizebox{\columnwidth}{!}{%
\begin{tabular}{ccccccccccc}
  \toprule
  \textbf{Retriever} & \textbf{Params} & \textbf{Metric} & \textbf{@1} & \textbf{@5} & \textbf{@10} & \textbf{@50} & \textbf{@100} & \textbf{@500} & \textbf{@1000} & \textbf{ms/query} \\
  \midrule
    \multirow{3}{*}{bge-base}                             & \multirow{3}{*}{110M} & R   & \textbf{0.235} & \textbf{0.415} & \textbf{0.477} & \textbf{0.606} & \textbf{0.641} & \textbf{0.803} & \textbf{0.843} & \multirow{3}{*}{22.8} \\
                                          &                            & P   & \textbf{0.471} & \textbf{0.207} & \textbf{0.121} & \textbf{0.032} & \textbf{0.017} & \textbf{0.004} & \textbf{0.002} & \\
                                          &                            & MRR & \multicolumn{7}{c}{\textbf{0.553}} & \\
    \midrule
    \multirow{3}{*}{bge-m3}                               & \multirow{3}{*}{568M} & R   & 0.199 & 0.361 & 0.420 & 0.531 & 0.594 & 0.741 & 0.806 & \multirow{3}{*}{46.3} \\
                                          &                            & P   & 0.391 & 0.166 & 0.105 & 0.027 & 0.015 & 0.004 & 0.002 & \\
                                          &                            & MRR & \multicolumn{7}{c}{0.479} & \\
    \midrule
    \multirow{3}{*}{e5-base}                              & \multirow{3}{*}{110M} & R   & 0.185 & 0.298 & 0.353 & 0.511 & 0.548 & 0.698 & 0.752 & \multirow{3}{*}{39.8} \\
                                          &                            & P   & 0.379 & 0.145 & 0.086 & 0.027 & 0.015 & 0.004 & 0.002 & \\
                                          &                            & MRR & \multicolumn{7}{c}{0.449} & \\
    \midrule
    \multirow{3}{*}{bge-m3 + content}                     & \multirow{3}{*}{568M} & R   & 0.123 & 0.299 & 0.366 & 0.463 & 0.507 & 0.655 & 0.730 & \multirow{3}{*}{47.3} \\
                                          &                            & P   & 0.241 & 0.133 & 0.089 & 0.023 & 0.012 & 0.003 & 0.002 & \\
                                          &                            & MRR & \multicolumn{7}{c}{0.358} & \\
    \midrule
    \multirow{3}{*}{bge-base + content}                   & \multirow{3}{*}{110M} & R   & 0.102 & 0.302 & 0.362 & 0.465 & 0.503 & 0.618 & 0.715 & \multirow{3}{*}{\textbf{22.1}} \\
                                          &                            & P   & 0.172 & 0.133 & 0.085 & 0.024 & 0.013 & 0.003 & 0.002 & \\
                                          &                            & MRR & \multicolumn{7}{c}{0.304} & \\
    \midrule
    \multirow{3}{*}{bm25}                                 & \multirow{3}{*}{---} & R   & 0.111 & 0.211 & 0.238 & 0.290 & 0.348 & 0.459 & 0.496 & \multirow{3}{*}{1592.1} \\
                                          &                           & P   & 0.195 & 0.097 & 0.055 & 0.015 & 0.009 & 0.002 & 0.001 & \\
                                          &                           & MRR & \multicolumn{7}{c}{0.266} & \\
  \bottomrule
\end{tabular}%
}

%% file: tables/9_retriever_query_config.tex
\resizebox{\columnwidth}{!}{%
\begin{tabular}{cccccccccccc}
  \toprule
  \textbf{Agg.} & \textbf{$M$} & \textbf{$tk$} & \textbf{R@1} & \textbf{R@5} & \textbf{R@10} & \textbf{R@50} & \textbf{R@100} & \textbf{R@500} & \textbf{R@1000} & \textbf{MRR} & \textbf{ms/task} \\
  \midrule
    ---   & 1 & ---  & \textbf{0.209} & \textbf{0.417} & \textbf{0.491} & 0.596 & 0.680 & 0.821 & 0.865 & \textbf{0.534} & \textbf{21.5} \\
    rrf   & 3 & ---  & 0.137 & 0.286 & 0.385 & 0.549 & 0.655 & 0.846 & 0.884 & 0.382 & 1597.5 \\
    rrf   & 5 & ---  & 0.146 & 0.292 & 0.407 & 0.586 & 0.664 & 0.837 & 0.891 & 0.384 & 2100.4 \\
    union & 3 & 100  & 0.165 & 0.363 & 0.433 & \textbf{0.606} & \textbf{0.687} & 0.767 & 0.767 & 0.464 & 931.3 \\
    union & 3 & 200  & 0.165 & 0.363 & 0.433 & 0.606 & 0.687 & 0.811 & 0.813 & 0.464 & 931.3 \\
    union & 3 & 500  & 0.165 & 0.363 & 0.433 & 0.606 & 0.687 & 0.814 & 0.847 & 0.464 & 931.3 \\
    union & 5 & 100  & 0.174 & 0.376 & 0.469 & 0.603 & 0.670 & 0.831 & 0.831 & 0.486 & 1945.1 \\
    union & 5 & 200  & 0.174 & 0.376 & 0.469 & 0.603 & 0.670 & \textbf{0.859} & \textbf{0.905} & 0.487 & 1945.1 \\
    union & 5 & 500  & 0.174 & 0.376 & 0.469 & 0.603 & 0.670 & 0.838 & 0.891 & 0.487 & 1945.1 \\
  \bottomrule
\end{tabular}%
}

%% file: tables/octen_ceiling.tex
\resizebox{\columnwidth}{!}{%
\begin{tabular}{lccccc}
  \toprule
  \textbf{Retriever (params)} & \textbf{R@5} & \textbf{R@10} & \textbf{R@100} & \textbf{R@1000} & \textbf{MRR} \\
  \midrule
  \texttt{BAAI/bge-base-en-v1.5} (110M, deployed) & 0.376 & 0.469 & 0.670 & 0.905 & 0.487 \\
  \texttt{Octen/Octen-Embedding-8B} (8B) & \textbf{0.410} & \textbf{0.549} & \textbf{0.817} & \textbf{0.941} & \textbf{0.518} \\
  \bottomrule
\end{tabular}%
}

%% file: tables/structure_aware.tex
\resizebox{\columnwidth}{!}{%
\begin{tabular}{lccccc}
  \toprule
  \textbf{Variant} & \textbf{MRR} & \textbf{R@5} & \textbf{R@10} & \textbf{R@100} & \textbf{R@1000} \\
  \midrule
  desc-only (bge-base, baseline) & \textbf{0.553} & 0.415 & 0.477 & 0.641 & 0.843 \\
  C2: desc + bge-code-v1 & 0.229 & 0.136 & 0.171 & 0.255 & 0.378 \\
  C1: content + bge-code-v1 & 0.097 & 0.085 & 0.092 & 0.126 & 0.226 \\
  A: 3-way (bge-base) · RRF & 0.500 & 0.426 & 0.527 & 0.754 & \textbf{0.925} \\
  A: 3-way (bge-base) · max & 0.535 & 0.467 & \textbf{0.556} & 0.740 & 0.914 \\
  A: 3-way (bge-base) · mean & 0.522 & 0.443 & 0.511 & 0.691 & 0.922 \\
  A: 3-way (bge-base) · sum-norm & 0.531 & 0.470 & 0.554 & \textbf{0.773} & 0.919 \\
  B: 3-way (mixed) · RRF & 0.506 & 0.464 & 0.531 & 0.709 & 0.906 \\
  B: 3-way (mixed) · max & 0.501 & 0.431 & 0.502 & 0.696 & 0.895 \\
  B: 3-way (mixed) · mean & 0.172 & 0.069 & 0.162 & 0.714 & 0.909 \\
  B: 3-way (mixed) · sum-norm & 0.528 & \textbf{0.476} & 0.534 & 0.704 & 0.896 \\
  \bottomrule
\end{tabular}%
}

%% file: tables/oracle_query_gap.tex
\begin{tabular}{lcc}
  \toprule
  \textbf{Rank bucket} & \textbf{\# oracle skills} & \textbf{Out-of-query} \\
  \midrule
  Top-5 (rank 1--5)                     & 97 & 4\% \\
  Top-100 (rank 6--100)                 & 72 & 8\% \\
  Worst-ranked (rank $>$ 100 or missing) & 60 & 48\% \\
  \bottomrule
\end{tabular}

%% file: tables/k1_sweep.tex
\begin{tabular}{cc}
  \toprule
  \textbf{Stage-1 cutoff $k_1$} & \textbf{Recall@$k_1$ (cascade ceiling)} \\
  \midrule
  100  & 0.670 \\
  250  & 0.793 \\
  500  & 0.838 \\
  1000 & 0.892 \\
  2000 & 0.932 \\
  4000 & 0.932 \\
  \bottomrule
\end{tabular}

%% file: tables/10_reranker_comparison.tex
\resizebox{\columnwidth}{!}{%
\begin{tabular}{cccccccccc}
  \toprule
  \textbf{Model} & \textbf{Agg.} & \textbf{$M$} & \textbf{MRR} & \textbf{R@1} & \textbf{R@5} & \textbf{R@10} & \textbf{R@50} & \textbf{R@100} & \textbf{ms/task} \\
  \midrule
    BGE-reranker-v2-m3 & ---  & 1 & \textbf{0.649} & \textbf{0.276} & \textbf{0.465} & \textbf{0.557} & \textbf{0.720} & \textbf{0.786} & 11999.3 \\
    \midrule
    \multirow{7}{*}{ms-marco-MiniLM} & \underline{---}  & \underline{1} & \underline{0.587} & \underline{0.259} & \underline{0.460} & \underline{0.520} & \underline{0.687} & \underline{0.776} & \underline{\textbf{1135.2}} \\
     & max  & 3 & 0.438 & 0.148 & 0.333 & 0.438 & 0.634 & 0.687 & 3024.7 \\
     & max  & 5 & 0.433 & 0.129 & 0.354 & 0.436 & 0.651 & 0.759 & 5098.8 \\
     & mean  & 3 & 0.485 & 0.177 & 0.387 & 0.465 & 0.639 & 0.720 & 3024.7 \\
     & mean  & 5 & 0.518 & 0.191 & 0.400 & 0.471 & 0.659 & 0.737 & 5098.8 \\
     & rrf  & 3 & 0.449 & 0.169 & 0.324 & 0.416 & 0.643 & 0.754 & 3024.7 \\
     & rrf  & 5 & 0.429 & 0.162 & 0.352 & 0.414 & 0.664 & 0.762 & 5098.8 \\
  \bottomrule
\end{tabular}%
}

%% file: tables/reranker_modern.tex
\resizebox{\columnwidth}{!}{%
\begin{tabular}{lccccccc}
  \toprule
  \textbf{Reranker (params)} & \textbf{Max Chars} & \textbf{R@5} & \textbf{R@10} & \textbf{R@100} & \textbf{R@1000} & \textbf{MRR} & \textbf{P@1} \\
  \midrule
  \texttt{cross-encoder/ms-marco-MiniLM-L-6-v2} (22.7M, deployed) & 512 & 0.460 & 0.520 & 0.776 & 0.905 & 0.587 & 0.494 \\
  \texttt{BAAI/bge-reranker-v2-m3} (568M) & 512 & \textbf{0.465} & \textbf{0.557} & \textbf{0.786} & 0.905 & \textbf{0.649} & \textbf{0.563} \\
  \texttt{Qwen/Qwen3-Reranker-0.6B} (596M) & 512 & 0.420 & 0.552 & 0.777 & 0.905 & 0.541 & 0.414 \\
  \texttt{Qwen/Qwen3-Reranker-0.6B} (596M) & 1024 & 0.451 & 0.530 & 0.766 & 0.905 & 0.629 & \textbf{0.563} \\
  \bottomrule
\end{tabular}%
}

%% file: tables/11_deep_reranker_config.tex
\resizebox{\columnwidth}{!}{%
\renewcommand{\arraystretch}{0.82}%
\begin{tabular}{cccccccc}
  \toprule
  \textbf{Chunk Size} & \textbf{Agg.} & \textbf{$M$} & \textbf{MRR} & \textbf{R@1} & \textbf{R@5} & \textbf{R@10} & \textbf{ms/task} \\
  \midrule
    1024 & --- & 1 & \textbf{0.653} & 0.270 & 0.476 & 0.568 & \textbf{2988.7} \\
    1024 & max & 3 & 0.517 & 0.166 & 0.420 & 0.537 & 7894.1 \\
    1024 & max & 5 & 0.552 & 0.197 & 0.439 & 0.536 & 27875.1 \\
    1024 & mean & 3 & 0.558 & 0.198 & 0.429 & 0.545 & 7894.1 \\
    1024 & mean & 5 & 0.573 & 0.208 & 0.472 & 0.539 & 27875.1 \\
    1024 & rrf & 3 & 0.505 & 0.194 & 0.413 & 0.519 & 7894.1 \\
    1024 & rrf & 5 & 0.510 & 0.185 & 0.418 & 0.512 & 27875.1 \\
    \addlinespace[2pt]
    2048 & --- & 1 & 0.637 & \textbf{0.272} & 0.490 & 0.585 & 5937.5 \\
    2048 & max & 3 & 0.521 & 0.175 & 0.418 & 0.525 & 18379.1 \\
    2048 & max & 5 & 0.535 & 0.183 & 0.433 & 0.518 & 29338.4 \\
    2048 & mean & 3 & 0.557 & 0.219 & 0.460 & 0.556 & 18379.1 \\
    2048 & mean & 5 & 0.580 & 0.232 & 0.447 & 0.545 & 29338.4 \\
    2048 & rrf & 3 & 0.478 & 0.164 & 0.389 & 0.530 & 18379.1 \\
    2048 & rrf & 5 & 0.466 & 0.147 & 0.408 & 0.505 & 29338.4 \\
    \addlinespace[2pt]
    4096 & --- & 1 & 0.634 & 0.271 & \textbf{0.499} & \textbf{0.595} & 12773.9 \\
    4096 & max & 3 & 0.522 & 0.172 & 0.443 & 0.544 & 37757.2 \\
    4096 & max & 5 & 0.529 & 0.183 & 0.412 & 0.541 & 129426.4 \\
    4096 & mean & 3 & 0.532 & 0.205 & 0.463 & 0.537 & 37757.2 \\
    4096 & mean & 5 & 0.587 & 0.246 & 0.431 & 0.566 & 129426.4 \\
    4096 & rrf & 3 & 0.458 & 0.159 & 0.380 & 0.526 & 37757.2 \\
    4096 & rrf & 5 & 0.489 & 0.197 & 0.400 & 0.532 & 129426.4 \\
    \addlinespace[2pt]
    8192 & --- & 1 & 0.620 & 0.251 & 0.467 & 0.553 & 30462.3 \\
    8192 & max & 3 & 0.509 & 0.169 & 0.402 & 0.552 & 95319.6 \\
    8192 & max & 5 & 0.506 & 0.178 & 0.401 & 0.523 & 154890.4 \\
    8192 & mean & 3 & 0.534 & 0.220 & 0.421 & 0.530 & 95319.6 \\
    8192 & mean & 5 & 0.547 & 0.227 & 0.425 & 0.530 & 154890.4 \\
    8192 & rrf & 3 & 0.468 & 0.179 & 0.392 & 0.492 & 95319.6 \\
    8192 & rrf & 5 & 0.467 & 0.181 & 0.400 & 0.475 & 154890.4 \\
  \bottomrule
\end{tabular}%
}

%% file: tables/20_selector_ablation.tex
\begin{tabular}{lcc}
  \toprule
  \textbf{Pipeline} & \textbf{Pass@1} & \textbf{Cost/Task} \\
  \midrule
  No Skills & 9.2{\scriptsize~[4.1, 14.9]} & \$0.026 \\
  SkillFlow (with selector) & 16.4{\scriptsize~[9.7, 23.6]} & \$0.025 \\
  Deep-reranker top-10 (no selector) & 15.9{\scriptsize~[9.7, 23.1]} & \$0.026 \\
  \bottomrule
\end{tabular}

%% file: tables/12_excluded_tasks.tex
\resizebox{\columnwidth}{!}{%
\begin{tabular}{llll}
  \toprule
  \textbf{Benchmark} & \textbf{Task} & \textbf{Category} & \textbf{Reason} \\
  \midrule
  SkillsBench & earthquake-phase-association   & Infrastructure & Daytona sandbox error \\
  SkillsBench & fix-druid-loophole-cve         & Infrastructure & Daytona sandbox error \\
  SkillsBench & fix-erlang-ssh-cve             & Infrastructure & Daytona sandbox error \\
  SkillsBench & latex-formula-extraction       & Infrastructure & Daytona sandbox error \\
  SkillsBench & organize-messy-files           & Infrastructure & Daytona sandbox error \\
  SkillsBench & parallel-tfidf-search          & Infrastructure & Daytona sandbox error \\
  SkillsBench & quantum-numerical-simulation   & Infrastructure & Daytona sandbox error \\
  SkillsBench & setup-fuzzing-py               & Infrastructure & Daytona sandbox error \\
  SkillsBench & shock-analysis-demand          & Infrastructure & Daytona sandbox error \\
  SkillsBench & syzkaller-ppdev-syzlang        & Infrastructure & Daytona sandbox error \\
  SkillsBench & taxonomy-tree-merge            & Infrastructure & Daytona sandbox error \\
  SkillsBench & video-filler-word-remover      & Infrastructure & Daytona sandbox error \\
  SkillsBench & video-tutorial-indexer         & Infrastructure & Daytona sandbox error \\
  SkillsBench & multilingual-video-dubbing     & Infrastructure & TTS model download fails intermittently during Docker build \\
  SkillsBench & scheduling-email-assistant     & Infrastructure & Docker compose mounts hardcoded host path \\
  SkillsBench & speaker-diarization-subtitles  & Infrastructure & Whisper model loading triggers OOM during Docker build \\
  \midrule
  SkillsBench & dynamic-object-aware-egomotion  & Reproducibility & Oracle outputs non-serializable numpy types \\
  SkillsBench & fix-build-google-auto           & Reproducibility & Maven build flaky under Docker networking \\
  SkillsBench & pedestrian-traffic-counting     & Reproducibility & Oracle depends on external vision API keys \\
  SkillsBench & r2r-mpc-control                 & Reproducibility & MPC settling time sensitive to Docker CPU scheduling \\
  SkillsBench & reserves-at-risk-calc           & Reproducibility & Numerical precision varies across platforms \\
  SkillsBench & simpo-code-reproduction        & Reproducibility & Build timeout; passes with extended timeout \\
  \midrule
  Terminal-Bench & train-fasttext               & Resource      & Large corpus download and model training \\
  \bottomrule
\end{tabular}%
}

%% file: tables/13_case_study_1.tex
\resizebox{\columnwidth}{!}{%
\begin{tabular}{llrl}
  \toprule
  \textbf{Stage} & \textbf{Skill} & \textbf{Score} & \textbf{GT?} \\
  \midrule
  Stage 1: Dense Retriever & \texttt{pdf-extractor-1} & 0.811 &  \\
   & \texttt{pdfco} & 0.766 &  \\
   & \texttt{llmwhisperer} & 0.748 &  \\
   & \texttt{pandas-best-practices} & 0.746 &  \\
   & \texttt{processing-data} & 0.745 &  \\
  \midrule
  Stage 2: Cross-Encoder & \texttt{fuzzy-matching} & 3.743 &  \\
   & \texttt{transaction-classification-debugger} & 1.156 &  \\
   & \texttt{fuzzy-match} & 0.025 & \checkmark \\
   & \texttt{fuzzy-match} & 0.025 &  \\
   & \texttt{nucleo-matcher} & -1.767 &  \\
  \midrule
  Stage 3: Deep Reranker & \texttt{fuzzy-matching} & 0.911 &  \\
   & \texttt{fuzzy-match} & 0.829 & \checkmark \\
   & \texttt{fuzzy-match} & 0.829 &  \\
   & \texttt{transaction-classification-debugger} & 0.600 &  \\
   & \texttt{nucleo-matcher} & 0.285 &  \\
  \midrule
  Stage 4: LLM Selector & \texttt{fuzzy-matching} & relevant &  \\
   & \texttt{fuzzy-match} & filtered & \checkmark \\
   & \texttt{fuzzy-match} & filtered &  \\
   & \texttt{transaction-classification-debugger} & filtered &  \\
   & \texttt{nucleo-matcher} & filtered &  \\
  \bottomrule
\end{tabular}%
}

%% file: tables/14_case_study_2.tex
\resizebox{\columnwidth}{!}{%
\begin{tabular}{llrl}
  \toprule
  \textbf{Stage} & \textbf{Skill} & \textbf{Score} & \textbf{GT?} \\
  \midrule
  Stage 1: Dense Retriever & \texttt{pdf-4} & 0.761 &  \\
   & \texttt{python-pdf} & 0.754 &  \\
   & \texttt{writing-tests-4} & 0.749 &  \\
   & \texttt{fixture-table} & 0.745 &  \\
   & \texttt{pdf-19} & 0.740 &  \\
  \midrule
  Stage 2: Cross-Encoder & \texttt{pdf} & -0.736 &  \\
   & \texttt{pdf-33} & -0.736 &  \\
   & \texttt{pdf-30} & -0.736 &  \\
   & \texttt{pdf-28} & -0.736 &  \\
   & \texttt{pdf-26} & -0.987 &  \\
  \midrule
  Stage 3: Deep Reranker & \texttt{python-pdf} & 0.706 &  \\
   & \texttt{pdf-16} & 0.698 &  \\
   & \texttt{pdf-20} & 0.687 &  \\
   & \texttt{pdf-7} & 0.683 &  \\
   & \texttt{pdf-21} & 0.683 &  \\
  \midrule
  Stage 4: LLM Selector & \texttt{python-pdf} & relevant &  \\
   & \texttt{pdf-16} & filtered &  \\
   & \texttt{pdf-20} & filtered &  \\
   & \texttt{pdf-7} & filtered &  \\
   & \texttt{pdf-21} & filtered &  \\
  \bottomrule
\end{tabular}%
}

%% file: tables/15_case_study_3.tex
\resizebox{\columnwidth}{!}{%
\begin{tabular}{llrl}
  \toprule
  \textbf{Stage} & \textbf{Skill} & \textbf{Score} & \textbf{GT?} \\
  \midrule
  Stage 1: Dense Retriever & \texttt{matched-filtering} & 0.844 & \checkmark \\
   & \texttt{conditioning} & 0.757 & \checkmark \\
   & \texttt{gwosc} & 0.694 &  \\
   & \texttt{scipy-best-practices} & 0.656 &  \\
   & \texttt{tuning-hyperparameters} & 0.650 &  \\
  \midrule
  Stage 2: Cross-Encoder & \texttt{matched-filtering} & 7.510 & \checkmark \\
   & \texttt{conditioning} & 4.322 & \checkmark \\
   & \texttt{gwosc} & -4.145 &  \\
   & \texttt{large-cell-ratio-matching} & -5.281 &  \\
   & \texttt{detect-python-command} & -6.234 &  \\
  \midrule
  Stage 3: Deep Reranker & \texttt{matched-filtering} & 0.999 & \checkmark \\
   & \texttt{conditioning} & 0.989 & \checkmark \\
   & \texttt{gwosc} & 0.270 &  \\
   & \texttt{bio-filter-sequences} & 0.156 &  \\
   & \texttt{oe-keyword-audit} & 0.094 &  \\
  \midrule
  Stage 4: LLM Selector & \texttt{matched-filtering} & filtered & \checkmark \\
   & \texttt{conditioning} & filtered & \checkmark \\
   & \texttt{gwosc} & filtered &  \\
   & \texttt{bio-filter-sequences} & filtered &  \\
   & \texttt{oe-keyword-audit} & filtered &  \\
  \bottomrule
\end{tabular}%
}

%% file: tables/18_skill_usage.tex
\resizebox{\columnwidth}{!}{%
\begin{tabular}{lccccc}
  \toprule
  \textbf{Injected} & \textbf{Tasks} & \textbf{SkillFlow} & \textbf{Baseline} & \textbf{Helped} & \textbf{Hurt} \\
  \midrule
  All tasks & 64 & 0.167 & 0.094 & 16 & 4 \\
  Oracle present & 55 & 0.194 & 0.103 & 16 & 3 \\
  Non-oracle only & 4 & 0.000 & 0.000 & 0 & 0 \\
  No skills & 5 & 0.000 & 0.067 & 0 & 1 \\
  \bottomrule
\end{tabular}%
}